\pdfoutput=1
\PassOptionsToPackage{pagebackref,breaklinks,colorlinks,allcolors=forestgreen}{hyperref}
\documentclass[11pt]{article}

\usepackage[final]{acl}

\usepackage{times}
\usepackage{latexsym}
\usepackage[T1]{fontenc}
\usepackage[utf8]{inputenc}
\usepackage{microtype}

\usepackage{inconsolata}
\usepackage{graphicx}
\usepackage[utf8]{inputenc} 
\usepackage[T1]{fontenc}    
\usepackage{url}            
\usepackage{booktabs}       
\usepackage{amsfonts}       
\usepackage{nicefrac}       
\usepackage{microtype}      
\usepackage{xcolor}         
\usepackage{wrapfig}
\usepackage{subcaption}

\usepackage{graphicx}
\usepackage{subcaption}
\usepackage{enumitem}
\usepackage{pgfplots}
\pgfplotsset{compat=1.17}
\usepackage{tikz}
\usepackage{booktabs}
\usepackage{colortbl}
\usepackage{xcolor}

\definecolor{forestgreen}{RGB}{34, 139, 34}
\newcommand{\name}{LexiCLIP}

\newcommand{\lexismall}{\textbf{\name{} (0.3B)}}

\newcommand{\lexift}{\textbf{\name{} --\textsc{ft}}}
\newcommand{\lexizs}{\textbf{\name{} --\textsc{zs}}}

\newcommand{\lexismallzs}{\textbf{\name{} (0.3B)--\textsc{zs}}}
\newcommand{\lexismallft}{\textbf{\name{} (0.3B)--\textsc{ft}}}
\newcommand{\lexilargezs}{\textbf{\name{} (7B)--\textsc{zs}}}

\newcommand{\lexismallinline}{\name{} (0.3B)}

\title{Vision-Free Retrieval: Rethinking Multimodal Search with Textual Scene Descriptions}

\author{
\begin{tabular}{c}
  Ioanna Ntinou$^{1}$\thanks{Equal contribution.} \quad
  Alexandros Xenos$^{1}$\footnotemark[1] \quad
  Yassine Ouali$^{2}$ \\
  Adrian Bulat$^{2,3}$ \quad
  Georgios Tzimiropoulos$^{1,2}$
\end{tabular} \\
  $^{1}$Queen Mary University of London, UK \\
  $^{2}$Samsung AI Centre, Cambridge, UK \\
  $^{3}$Technical University of Iași, Romania \\
  \texttt{\{i.ntinou,a.xenos,g.tzimiropoulos\}@qmul.ac.uk} \\
  \texttt{y.ouali@samsung.com \quad adrian@adrianbulat.com}
}

\begin{document}
\maketitle
\begin{abstract}
Contrastively-trained Vision-Language Models (VLMs), such as CLIP, have become the standard approach for learning discriminative vision-language representations. However, these models often exhibit shallow language understanding, manifesting bag-of-words behaviour. These limitations are reinforced by their dual-encoder design, which induces a \textit{modality gap}. Additionally, the reliance on vast web-collected data corpora for training makes the process computationally expensive and introduces significant privacy concerns. To address these limitations, in this work, we challenge the necessity of vision encoders for retrieval tasks by introducing a \textit{vision-free, single-encoder} retrieval pipeline. Departing from the traditional text-to-image retrieval paradigm, we migrate to a text-to-text paradigm with the assistance of VLLM-generated structured image descriptions.   We demonstrate that this paradigm shift has significant advantages, including a substantial reduction of the modality gap, improved compositionality, and better performance on short and long caption queries, all attainable with only a few hours of calibration on two GPUs. Additionally, substituting raw images with textual descriptions introduces a more privacy-friendly alternative for retrieval. To further assess generalisation and address some of the shortcomings of prior compositionality benchmarks, we release two benchmarks derived from Flickr30k and COCO, containing diverse compositional queries made of short captions, which we coin subFlickr and subCOCO. Our vision-free retriever matches and often surpasses traditional multimodal models. Importantly, our approach achieves state-of-the-art zero-shot performance on multiple retrieval and compositionality benchmarks, with models as small as 0.3B parameters. Code is available at \href{https://github.com/IoannaNti/LexiCLIP}{LexiCLIP}.

\end{abstract}

\section{Introduction}

Contrastively-trained Vision-Language Models (VLMs)~\cite{pmlr-v139-radford21a} have rapidly become a cornerstone for learning powerful, discriminative vision-language representations. Their success is underscored by remarkable zero-shot transfer abilities across a wide array of tasks \cite{pmlr-v139-jia21b, pmlr-v162-li22n, li2022supervisionexistseverywheredata, pmlr-v139-radford21a, Zhai_2023_ICCV}. These capabilities are largely attributed to their training on vast quantities of image-text pairs using a simple contrastive objective. However, this scale comes at a significant cost: training such models is computationally expensive, and the reliance on web-collected data introduces notable privacy challenges. Moreover, the prevalent dual-tower architecture that encodes images and text separately induces a \textit{modality gap}, an effect which hinders the model's fairness and compositional abilities~\cite{liang2022mind}. The latter also stems from the limited understanding of language structure of the CLIP's text encoder, whose representations tend to ignore word order and syntactic relations – effectively treating the caption as an unordered bag-of-words \cite{yuksekgonul2023visionlanguagemodelsbehavelike}. Due to the above limitations, contrastively trained VLMs (e.g., CLIP, SigLIP~\cite{zhai2023sigmoid}) often exhibit poor compositional generalization and shallow language understanding, yet they achieve strong performance on popular text-image retrieval benchmarks like Flickr30k~\cite{10.1162/tacl_a_00166} and COCO~\cite{10.1007/978-3-319-10602-1_48}.

A growing body of research has quantified the limitations of current vision-language models and begun to address them through several approaches. These include constructing or mining hard negative examples~\cite{yuksekgonul2023visionlanguagemodelsbehavelike}, employing shared or partially shared backbones~\cite{likhosherstov2021polyvit}, and adding cross-modal fusion modules or adapters to learn fine-grained alignment between image regions and words~\cite{li2023blip2}. However, the use of hard negatives has been shown to potentially rely on shortcuts or spurious patterns~\cite{NEURIPS2023_63461de0} while approaches based on cross-modal fusion are impractical due to the need for a separate inference pass for each image with every new query.

Departing from previous works, we aim to (1) remove the modality gap by design, (2) reduce the \textit{bag-of-words} behavior, and (3) alleviate the privacy concerns pertaining to the training data; all under a framework that requires limited training and data. Finally, (4) we seek to introduce a new text-image retrieval benchmark that cannot be easily solved by VLMs exhibiting bag-of-words behaviour.

To this end, we propose a paradigm shift by converting images entirely into carefully crafted textual descriptions, thereby enabling language models to reason about visual content purely through text. This strategy offers significant advantages, including leveraging high-capacity pretrained text encoders, significantly narrowing the modality gap through a fully shared encoder, and substantial mitigation of privacy risks as the model avoids direct handling of sensitive image data. However, this approach faces a fundamental challenge: faithfully representing rich visual information solely with text remains an open and under-explored problem. To address this, we first introduce a robust, principled, and carefully designed pipeline for image-to-text conversion that captures the richness of visual information. We then show that the resulting text-based image representation can produce strong, vision-free, zero-shot text-image retrieval models.
To further boost the accuracy of the model, we utilize the textual corpus generated by applying the proposed pipeline to 1.5M images from the OpenImages dataset \cite{Kuznetsova_2020} to fine-tune the model, better aligning it to the input distribution.

Finally, recognizing the aforementioned limitations of existing text-image retrieval benchmarks, we introduce two new datasets, subFlickr and subCOCO (derived from the Flickr and COCO datasets, respectively), specifically designed to assess performance on short compositional tags, an area poorly represented in previous test suites, where, as we show, the standard VLMs appear to struggle. In summary, our main contributions are:
\begin{itemize}[leftmargin=*]
\itemsep0em 

\item We introduce \name, a novel text-only Vision-Language framework that converts images into textual descriptions, enabling language models to process visual content. This inherently removes the modality gap, reduces the ``bag-of-words'' effect, and alleviates privacy concerns, all while requiring limited to no training and data.
\item A new principled and carefully designed pipeline for accurately converting rich visual information into text, with ample validation on a multitude of benchmarks.
\item We introduce two new datasets, subFlickr and subCOCO, specifically curated to evaluate VLMs on short compositional queries, an area previously underrepresented in benchmarks.
\item Using solely textual inputs, and no task-specialised data, we set a new state-of-the-art result on image-text composionality and image retrieval with long captions.
\end{itemize}

\section{Related work}
\label{gen_inst}
\subsection{Text-only training}

Recent methods propose to drop images from the training pipelines in an attempt to alleviate the modality gap. Knight~\cite{wang2023associationgenerationtextonlycaptioning} introduces a text-only captioning pipeline where image- or video-derived captions are used to build a text corpus for training a decoder with autoregressive loss. At inference, the k-nearest captions are retrieved and used as embeddings to a decoder. DeCap~\cite{li2023decapdecodingcliplatents} trains a lightweight language decoder purely on a large corpus of text embeddings generated from CLIP's~\cite{radford2021learning} text encoder. CLOSE~\cite{Gu_2023_ICCV} observes the low image-text cosine similarity and proposes a hyper-parameter-scaled noise injection method. IFCap~\cite{lee2024ifcapimagelikeretrievalfrequencybased} similarly injects noise into text embeddings to imitate image embeddings, improving the retrieval of semantically aligned captions. CLIPPO~\cite{Tschannen_2023_CVPR} shifts from the dual encoder paradigm by jointly processing images and text (where alt-text is rendered as an image) using a purely pixel-based model. The resulting image pair is encoded with a shared vision encoder and trained via contrastive loss. Different from previous works, our pipeline is more privacy-friendly, excels at long-form text retrieval—overcoming the fixed sequence-length constraints of pixel-based encoders—and enables extensive linguistic knowledge transfer via strong pre-trained language models.

\subsection{Datasets for text-to-image retrieval}

Text-image retrieval datasets typically fall into two categories: long-caption and short-tag datasets. 
The conventional approach, common in many prior works, uses long captions. Key benchmarks are Flickr30k~\cite{10.1162/tacl_a_00166} (31K images) and MS COCO~\cite{10.1007/978-3-319-10602-1_48} (330K images), each offering five crowdsourced full-sentence descriptions per image, averaging 10–13 tokens and describing the full scene. NoCaps~\cite{agrawal2019nocaps} expands this to 15K OpenImages-derived images with 166K human-written captions covering a broader range of categories. These datasets are characterized by rich, syntactic, sentence-level annotations describing the entire image and often averaging over 10 words per caption. 

The newer and second line of research uses short tag datasets with keyword-style annotations.  Tag2test~\cite{huang2023tag2text}, RAM~\cite{zhang2023recognize} and RAM++~\cite{huang2023open}
automatically extract a set of tags from existing captions or metadata, yielding large-scale image–tag pair corpora without manual labeling.  Each image is labeled with a collection of salient keywords (e.g. ``dog'', ``couch'', ``table'' for a living-room scene) rather than a full sentence, enumerating the contents without syntax. Such tag-based datasets are often an order of magnitude larger, on the order of millions of images drawn from web data and covering thousands of distinct tag categories, e.g.  ~3,400 categories in Tag2Text handles and ~4583 in RAM. These tags lack sentence structure but reflect real-world search behavior more closely, where users input short, compositional queries.

In this work, we introduce \textit{subCOCO} and \textit{subFlickr}, which are positioned between these two extremes regarding annotation granularity. Built from Flickr30K and COCO, they use sub-sentential phrases, shorter than full captions but semantically richer than flat tags. An overview of these datasets and others is presented in Tab.~\ref{tab:extended_dataset_comparison}.

\begin{table*}[ht]
\centering
\setlength{\tabcolsep}{5.5pt}
\renewcommand{\arraystretch}{0.99}
\caption{Overview of text-image retrieval evaluation datasets used in this study. * - denotes estimated statistics.}
\label{tab:extended_dataset_comparison}
\resizebox{0.8\linewidth}{!}{%
  \begin{tabular}{lccccc}
    \toprule
    \textbf{Dataset} & \textbf{\# Images} & \textbf{\# Queries} 
      & \textbf{Avg.\ Query Length} & \textbf{Captions per Image} & \textbf{Query Type} \\
    \midrule
    Flickr30K~\cite{10.1162/tacl_a_00166}       & 1\,K    & 5\,K    & 13.4  & 5   & Full Sentence           \\
    MS-COCO~\cite{10.1007/978-3-319-10602-1_48} & 5\,K    & 25\,K   & 10.4  & 5   & Full Sentence           \\
    NOCAPs~\cite{agrawal2019nocaps}             & 10\,K   & 106\,K  & 9--11  & 10--11\textsuperscript{*} & Full Sentence \\
    Conceptual Captions~\cite{sharma2018conceptual} 
      & 22.5\,K & 22.5\,K & 9.7   & 1   & Sentence Caption        \\
    Winoground~\cite{Thrush_2022_CVPR}          & 400   & 400   & 8.8   & 1   & Compositional (Paired)     \\
    SugarCrepe~\cite{NEURIPS2023_63461de0}      & 7.5\,K    & 7.5\,K   & 10\textsuperscript{*}  & 1   & Compositional (Paired) \\
    SugarCrepe++~\cite{NEURIPS2024_200661bf}    & 4.8\,K    & 9.5\,K   & 10\textsuperscript{*}  & 2   & Compositional (Paired) \\
    ADE20K~\cite{zhou2017scene}                 & 2\,K    & N/A   & 9.9   & N/A & Tag-based Queries        \\
    OpenImages~\cite{kuznetsova2020open}        & 125\,K  & N/A   & 1--5   & 8   & Tag-based Queries        \\
    \midrule
    subFlickr                                  &  935   & 280  & 4.5   & 6.0 & Compositional Caption \\
    subCOCO                                    & 4030 & 256  & 3.47   & 4.1 & Compositional Caption \\
    \bottomrule
  \end{tabular}%
}
\vspace{-0.4cm}
\end{table*}

\section{Vision-Free Constrastive Learning}\label{sec:method}

\begin{figure}
    \centering

    \begin{subfigure}[t]{0.45\textwidth}
        \centering
        \includegraphics[width=\linewidth]{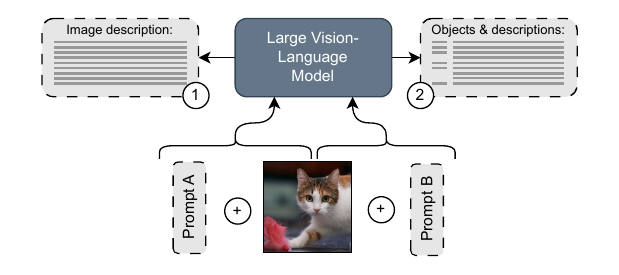}
        \caption{The proposed Image to Text conversion pipeline: An image is converted into an equivalent textual description in two steps: one prompted using ``Prompt A'' to describe the image in detail, and another, using ``Prompt B'' aimed at capturing each object and corresponding attributes.}
        \label{fig:text-to-image}
    \end{subfigure}%
    \hspace{0.02\textwidth}
    \begin{subfigure}[t]{0.36\textwidth}
        \centering
        \includegraphics[width=\linewidth]{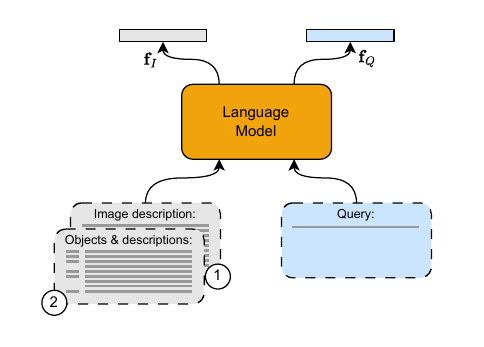}
        \caption{\name~- the proposed Vision-Free image-text retrieval model.}
        \vspace{-0.2cm}
        \label{fig:our-model}
    \end{subfigure}
    \caption{The proposed Vision-Free Retrieval Pipeline.} 
    \label{fig:overall_approach}
    \vspace{-0.4cm}
\end{figure}

Contrastive language-image pretraining has emerged as a highly effective method for developing vision-language models, leveraging vast amounts of web-collected image-text pairs. Using this data, the prevalent technique trains two independent encoders, one for each modality, using a contrastive loss that aims to map each input to a joint embedding space~\cite{radford2021learning}. Despite its success, this approach suffers from a series of drawbacks: (1) Training and finetuning such models is computationally expensive, (2) Using web collected images may result in privacy infringements and (3) The models suffer from a \textit{bag-of-words} behaviour~\cite{yuksekgonul2022and}, largely a consequence of the modality gap induced by the two separate towers~\cite{liang2022mind}.
As a solution to these issues, we introduce \name, a novel vision-free text-to-text contrastive learning framework that leverages pretrained language models for effective image-text retrieval within a shared single-tower architecture. Our key idea is to bring the images into the language domain via dense captioning, leveraging thereafter the world knowledge of discriminatively pretrained LLMs. The conversion to textual descriptions is also privacy-friendly, as most identity-related information (i.e., faces, private rooms, etc.) is removed.
Without any further training, in a zero-shot manner, our solution showcases strong image-text retrieval abilities, which we further boost using a light finetuning on a small dataset.

\begin{figure}[!htbp]
  \centering
  \includegraphics[
    width=0.4\textwidth,
    keepaspectratio,
    trim=10 10 10 10,
    clip
  ]{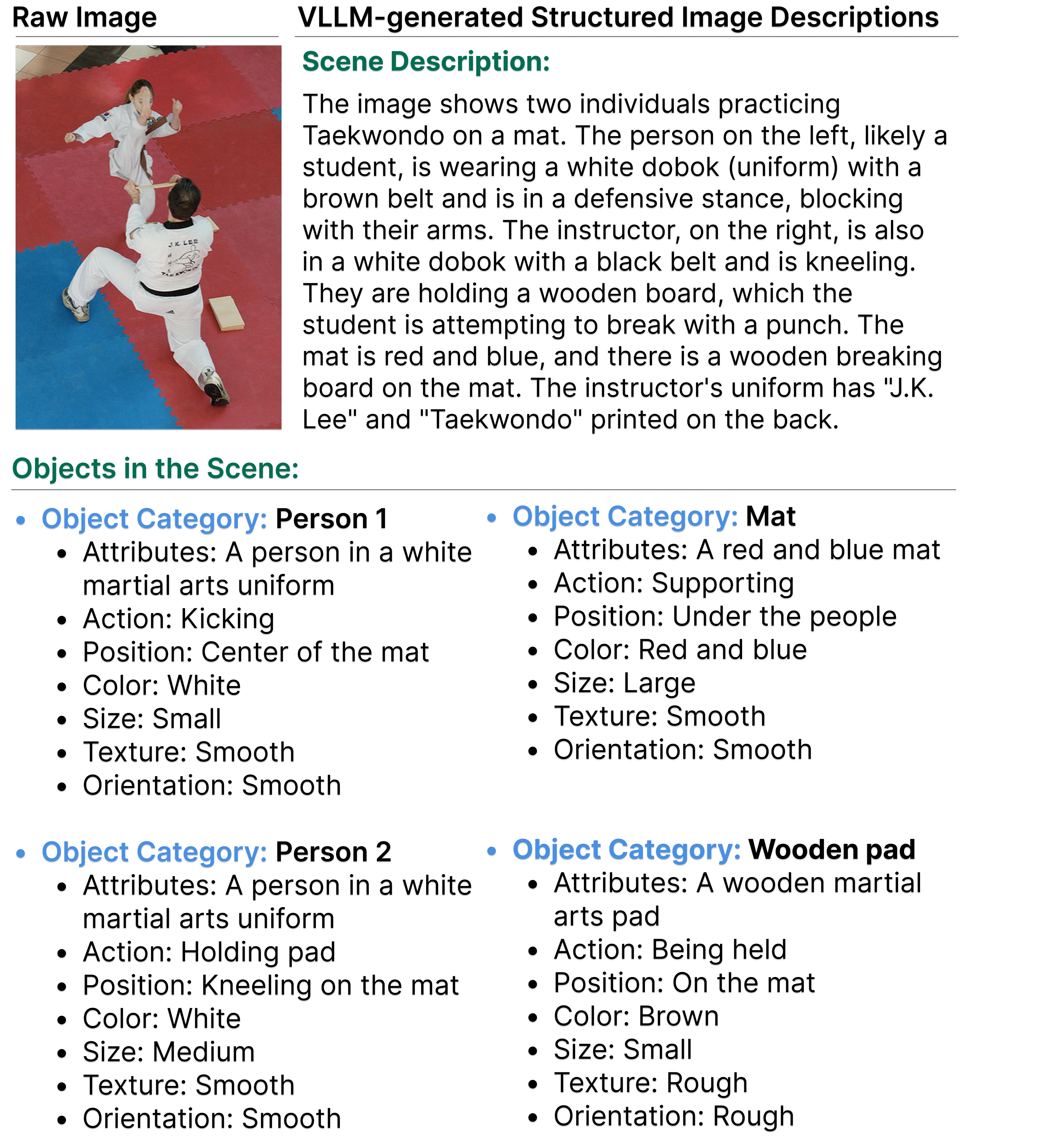}
  \caption{Example of our textual image representations.}
  \label{fig:image_rep_example}
  \vspace{-0.6cm}
\end{figure}

\noindent \textbf{Representing images using text:} \textit{A picture's worth: how many words, and which ones truly matter?} A wide disparity exists in the literature regarding the structure, content, and particularly the length of these textual representations, spanning from brief annotations of one word~\cite{gal2022image} to extensive descriptions comprising over a thousand words~\cite{collell2016image}. In the absence of a prevailing standard, we undertake an analysis of this question within the specific domain of image retrieval. We posit that a series of desirable properties characterize an effective textual descriptor: It should capture the (1) high-level scene details, identify (2) all salient objects, and articulate (3) actions, interactions, and object placement. Furthermore, its structure must be (4) coherent and (5) organized. 

Thanks to the rapid progress in the area of large vision language modeling~\cite{chen2024expanding,bai2025qwen2}, many of these requirements can be readily addressed by providing a Vision LLM (VLLM) with an image and an appropriate prompt that tasks the model to generate highly detailed image descriptions. Generally, we find that longer descriptions are desirable, as they better capture the richness of information present in an image. 

Nonetheless, this alone is insufficient. Due to significant variations in image object density, the model may occasionally fail to identify certain objects or inaccurately report or miss their attributes. To mitigate this issue, we propose a supplementary step utilizing the same VLLM: generating a structured list that enumerates the objects present in the image, accompanied by a brief description for each. Our findings indicate that a JSON format is optimal for structuring this output. 

We consider a wide variety of prompts and models. Out of the models tested, the best results were obtained using InternVL-2.5-8B-MPO ~\cite{wang2025enhancingreasoningabilitymultimodal} VLLM. We provide the final prompts used alongside a set of randomly sampled examples in the supplementary material. The overall image-to-text conversion process is shown in Fig.~\ref{fig:text-to-image}, while Fig.~\ref {fig:image_rep_example} shows an example of the resulting representation.

\noindent \textbf{Zero-shot vision-free image-text retrieval:} Given an image description $\mathcal{T}$ and a query $\mathcal{Q}$, the corresponding image and query embeddings, $\mathbf{f}_T$ and $\mathbf{f}_Q$ are obtained by passing each tokenized input through the shared language model $\Phi(\Theta,.)$. Following best practices for zero-shot evaluations,  we also concatenate a handcrafted prompt $p_Q$ for the query. The process is depicted in Fig.~\ref{fig:our-model}.

\noindent \textbf{Vision-Free text-to-text finetuning:} While single-tower language models exhibit robust zero-shot performance, finetuning presents opportunities for enhancement, especially in the case of smaller variants. Such improvements are motivated by two key factors: firstly, smaller models tend to have more constrained generalization, and secondly, the statistical distribution of image descriptions can diverge from the data typically used in the pretraining phase of these Language Models.

To this end, we construct a small alignment set by converting 1.5M images from the OpenImages dataset to textual representations using the aforementioned process. Since these images are not paired with short (query) captions, we synthetically generate them using VLLMs. In particular, we run  BLIP-2 to generate concise captions of around 10 tokens, and we prompt InternVL-2.5-8B-MPO ~\cite{wang2025enhancingreasoningabilitymultimodal} to extract a pool of six 2-3 word compositional captions.

For fine-tuning, we adopt a two-stage contrastive training paradigm. In the first stage, our model is trained to align longer BLIP-2 captions with images. As a second stage, the model is adapted using a mixture of short compositional captions and BLIP-2 captions, refining its understanding of fine-grained details. More details on the prompt and the training can be found in the appendix.

\section{subFlickr and subCOCO}

Most widely used image-text retrieval datasets, such as MS COCO and Flickr30K, feature full-sentence captions that average 10–13 words in length. These captions tend to be lengthy, formal, and grammatically complete. Moreover, benchmarks like MS COCO and Flickr30K  mostly rely on broad, scene-level descriptions, which do not reflect how people actually search for images. For example, a caption in Flickr30K might be ``A man in a green t-shirt and long tan apron hacks apart the carcass of a cow while another man hoses away the blood.'' but in real-world queries that could possibly be much shorter and more fragmented like ``man in a green t-shirt,'' or ``hack the carcass up.'' These natural queries are typically informal, ungrammatical, and omit function words, resembling spoken language rather than written prose.
On the other hand,  keyword-style annotations \cite{huang2023tag2text, zhang2023recognize, huang2023open} lack in semantic richness, compositionality, and alignment with real user queries. 

To address this limitation, we introduce two retrieval benchmarks, \textit{subFlickr} and \textit{subCOCO}, derived from the test sets of the Flickr30k and MS COCO datasets, respectively. To generate a set of concise queries, we decompose the existing ground-truth captions into shorter, meaningful subcaptions. For this, we employ a pretrained constituency parser~\cite{honnibal2020spacy} to decompose each caption to its constituent nodes, i.e. complete captions, sentences, sub-phrases, and individual lexical items (nouns, verbs, etc.). From this structure, we extract recurring subphrases that are likely to be visually grounded. We then manually curate a set of queries, choosing compositional expressions,  such as ``a person with a white shirt. These serve as retrieval queries in our benchmark. To match each image with the relevant queries,  we first compute text-to-text similarity scores between the ground truth subcaptions and the curated queries using a text encoder (Bge-large-en-v1.5~\cite{bge_embedding}). As a second step, we use two VLMs: Qwen2-VL-7B~\cite{bai2025qwen2} and InternVL2.5-8B~\cite{chen2024expanding} to verify whether each query is visually present in the image. A query is assigned to an image if both models agree, providing a more reliable and grounded labeling. As a final step, we visually inspect 20\% of our dataset. More details about our dataset can be found in the appendix. 

\section{Experiments}
\label{experimens}
We compare our approach with the current state-of-the-art in four tasks of interest: (1) compositional retrieval using short captions on our newly introduced benchmarks, (2) zero-shot text-to-image retrieval, (3) image-to-text long captions retrieval, and (4) compositional understanding. In each case, we compare our method against a broad set of state-of-the-art two-tower (independent) VLMs. For specialized tasks such as compositional understanding, we include targeted baselines where applicable.

We refer to our models evaluated in a zero-shot setting using structured text-only representations as \lexizs, and to their fine-tuned variants as \lexift.  We focus our evaluation on a 0.3B parameter encoder backbone configuration, denoted as \lexismall{}. Specifically, we adopt a robust instruction-tuned, contrastively pre-trained text model: BGE-large-en-v1.5~\cite{bge_embedding}, which achieves state-of-the-art performance on the MTE Benchmark~\cite{muennighoff2022mteb} within its model size. 

\subsection{Implementation details}
\label{ssec:implementation}
We train our model in two stages. For both stages, the model is trained for three epochs using the FlagEmbedding library~\cite{bge_embedding}, with a cosine-annealed learning rate schedule and a 5\% warm-up phase.  The model is trained on two A100 GPUs with an effective batch size of 2,048, a peak learning rate of $1\times 10^{-4}$.  
In the first stage, the training is done only using BLIP-2 captions.  In the second stage, we fine-tune the model using a mixture of BLIP-2 and compositional captions, where each batch of 2,048 samples consists of 200 concise BLIP-2 captions and the remainder compositional ones. The final model is obtained by averaging the checkpoints from the two stages, weighted 0.4 for the first and 0.6 for the second stage.

\subsection{Short caption retrieval}

We benchmark our datasets, \textit{subFlickr} and \textit{subCOCO}, over a series of models for short-caption text-to-image retrieval. The goal is to recognise all relevant images given a query, which is a short caption. Each query is associated with a binary relevance label over the test set. For evaluation, we opt for mean Average Precision (mAP) and F1-score.

Tab.~\ref{tab:map_f1_flickr_coco} reports results for both zero-shot and fine-tuned retrieval pipelines. In addition to a range of general two-tower models, we evaluate RAM~\cite{zhang2023recognize} and  RAM++~\cite{huang2023open} tagging models. We evaluate our method zero-shot, but also after fine-tuning. The latter, denoted as  \lexismallft{} in our table, and as proposed in this work, is a fine-tuned version of BGE-large-en-v1.5 that is trained exclusively on textual inputs—namely,  BLIP-2 captions and corresponding descriptive annotations - without using any image features during training (i.e. the text-only training introduced in Section~\ref{sec:method}).

 We note that both RAM~\cite{zhang2023recognize} and RAM++~\cite{huang2023open} achieve high zero-shot performance on our benchmarks. Their success is attributed to two key factors: first, these models are trained on a large-scale dataset of 14 million image-tag pairs, and second, they leverage explicit tag supervision, which allows them to learn fine-grained object-attribute associations.

\begin{table*}[!t]
\centering
\caption{Zero‐shot text–image retrieval metrics (mAP and F1@K) on SubFlickr and SubCOCO}
\tiny                               
\setlength{\tabcolsep}{1pt}      
\renewcommand{\arraystretch}{0.9} 
\resizebox{0.8\textwidth}{!}{%
  \begin{tabular}{@{}l c c c c   c c c c@{}}
    \toprule
    \textbf{Method} 
      & \multicolumn{4}{c}{\textbf{SubFlickr}} 
      & \multicolumn{4}{c}{\textbf{SubCOCO}} \\
    \cmidrule(lr){2-5} \cmidrule(lr){6-9}
      & mAP  & F1@1 & F1@5 & F1@10 
      & mAP  & F1@1 & F1@5 & F1@10 \\
    \midrule
    CLIP (ViT-B)~\cite{radford2021learning}    
      & 29.2 & 10.3 & 20.0 & 21.9
      & 33.6 & 4.5 & 11.9  & 17.5\\
    CLIP (ViT-L)~\cite{radford2021learning}    
      & 29.7 & 11.0 & 20.0 & 21.7
      & 36.1 & 4.8  & 13.4 & 19.5 \\
    OpenCLIP (ViT-G/14)~\cite{NEURIPS2022_a1859deb}          
      & 35.3 & 12.0 & 24.5 & 26.6
      & 41.2 & 6.3  & 15.9 & 22.0\\
    OpenCLIP (ViT-BigG/14)~\cite{NEURIPS2022_a1859deb}     
      & 36.5 & 13.3 & 25.4 & 26.8 
      & 42.1 & 6.3  & 16.1 & 22.6\\
    SigLIP ViT-B/16~\cite{zhai2023sigmoid}  
      & 36.6 & 13.6 & 25.6 & 27.2 
      &  43.6& 6.5 & 16.4& 23.0\\
EVA-02-CLIP (ViT-L-336)~\cite{fang2023eva02}        
      & 36.5 & 12.9 & 24.9 & 27.3 
      & 41.6 & 6.1  &15.3 & 21.6\\    
    RAM~\cite{zhang2023recognize}    
      & 48.3 & 14.7 & 32.2 & 34.2 
      & 50.8 &6.1 &17.2  & 26.0 \\
    RAM++~\cite{huang2023open}          
      & 49.1 &  15.3 &32.7 &  35.5
      & 52.5 &  6.2  &17.4 &  26.2\\
    \midrule
    \lexismallzs
      & 45.6 & 17.1 & 30.6 & 31.9
      & 48.3 & 6.2  & 17.1 & 25.4  \\
     \lexismallft{} & \textbf{55.1} & \textbf{17.9} &\textbf{37.6}  &\textbf{40.0}  & \textbf{54.3} & \textbf{6.7} & \textbf{18.5} & \textbf{27.5} \\
    \bottomrule
  \end{tabular}%
}
\label{tab:map_f1_flickr_coco}
\vspace{-0.3cm}
\end{table*}

\subsection{Zero-shot image-text retrieval}

\begin{table*}[!t]
\centering
\caption{Zero-shot text–image retrieval accuracy on Flickr30K and COCO.}
\tiny                               
\setlength{\tabcolsep}{1pt}           
\renewcommand{\arraystretch}{0.9} 
\resizebox{0.8\textwidth}{!}{%
  \begin{tabular}{@{}l c c c c   c c c c c@{}}
    \toprule
    \textbf{Method} 
      & \textbf{Params (B)} 
      & \multicolumn{4}{c}{\textbf{Image retrieval}} 
      & \multicolumn{4}{c}{\textbf{Text retrieval}} \\
    \cmidrule(lr){3-6} \cmidrule(lr){7-10}
      & 
      & \multicolumn{2}{c}{Flickr30K} 
      & \multicolumn{2}{c}{COCO}
      & \multicolumn{2}{c}{Flickr30K} 
      & \multicolumn{2}{c}{COCO} \\
      & 
      & R@1 & R@10 & R@1 & R@10 & R@1 & R@10 & R@1 & R@10 \\
    \midrule
    CLIP (ViT-B)~\cite{radford2021learning}            
      & 0.15 & 58.8 & 89.8 & 30.5 & 66.8 & 77.8 & 98.2 & 51.0 & 83.5 \\
    SigLIP ViT-B/16~\cite{zhai2023sigmoid} 
    & 0.23 & 74.6 & 95.6 & 47.8 & 81.0 & 89.1 & 99.3 & 65.7 & 91.3 \\
    CLIP (ViT-L)~\cite{radford2021learning}            
      & 0.43 & 67.3 & 93.3 & 37.0 & 71.5 & 87.2 & 99.4 & 58.1 & 87.8 \\
    BLIP (ViT-L)~\cite{li2022blip}                      
      & 0.23 & 70.0 & 95.2 & 48.4 & 83.2 & 75.5 & 97.7 & 63.5 & 92.5 \\
    BLIP2 (ViT-L)~\cite{li2023blip2}                   
      & 1.17 & 74.5 & 97.0 & 50.0 & 86.1 & 86.1 & 99.4 & 63.0 & \textbf{93.1} \\
    EVA-02-CLIP (ViT-L-336)~\cite{fang2023eva02}
      & 0.43 & 78.0 & 96.8 & 47.9 & 80.0 & 89.6 & 99.6 & 64.2 & 90.9 \\
    OpenCLIP (ViT-G/14)~\cite{NEURIPS2022_a1859deb}     
      & 1.37 & 77.8 & 96.9 & 48.8 & 81.5 & 91.5 & 99.6& 66.3 & 91.8 \\
    OpenCLIP (ViT-BigG/14)~\cite{NEURIPS2022_a1859deb} 
      & 2.54 & \textbf{79.5}& \textbf{97.5} & 51.3 & 83.0 & \textbf{92.9} & 97.1 & 67.3 & 92.6 \\
    \midrule
\lexismallzs & 0.3  & 69.5 & 94.2 & 41.7 & 76.7 & 75.9 & 97.4 & 45.4 & 80.3 \\
 \lexismallft{}   & 0.3  & 79.2 & 97.4 & \textbf{52.7} & \textbf{84.5} & 91.6 & \textbf{99.7} & \textbf{67.4} & 92.1  \\
    \bottomrule
  \end{tabular}
}
\label{table:flickr_coco_results}
\vspace{-0.5cm}
\end{table*}

We evaluate our approach on the standard Flickr30K \cite{10.1162/tacl_a_00166} and MSCOCO \cite{10.1007/978-3-319-10602-1_48} benchmarks. As Tab.~\ref{table:flickr_coco_results} shows, without any finetuning, our 300 M-parameter \name{} achieves 69.5 R@1 on Flickr30K and 41.7 R@1 on COCO, comparing favourably with similarly sized CLIP models. Post finetuning, our approach trained only on 1.5M text samples, matches and outperforms OpenCLIP (BiG/14), a 2.5B parameter model trained on 2B image-text pairs.

\subsection{Image-text long captions retrieval}

The CLIP model's ability to process longer text is greatly restricted by the text encoder, which typically can only process up to 77~\cite{radford2021learning} tokens. In practice, due to the data distribution of the captions, the effective length is even lower, at around 20-25 tokens. As our approach leverages pretrained language models trained on generic text, we posit that \name~ is well-suited for deployment for retrieval using long text. To test this, in Tab.~\ref{table:urban1k_results} we evaluate our approach on the Urban1k~\cite{Zhang2024LongCLIPUT} dataset. As the results show, without any finetuning, we already surpass (1) all other CLIP variants and (2) specialised CLIP models finetuned on long captions~\cite{Zhang2024LongCLIPUT}. 

With finetuning using the proposed approach, requiring no specialized data, we outperform prior state-of-the-art results by over 5\%.

\begin{table}[!ht]
  \centering
  \caption{Zero-shot text–image retrieval on Urban1k.}
  \small                           
  \setlength{\tabcolsep}{3pt}      
  \renewcommand{\arraystretch}{1.05}
  \resizebox{\columnwidth}{!}{%
    \begin{tabular}{@{}l c c c c@{}}
      \toprule
      \textbf{Method}
        & \multicolumn{2}{c}{\textbf{Image retrieval}}
        & \multicolumn{2}{c}{\textbf{Text retrieval}} \\
      \cmidrule(lr){2-3}\cmidrule(lr){4-5}
        & R@1 & R@10 & R@1 & R@10 \\
      \midrule
      CLIP (ViT-B)~\cite{radford2021learning}    
        & 46.5 & 78.7 & 62.5 & 90.5 \\
      SigLIP ViT-B/16~\cite{zhai2023sigmoid}
       & 62.1 & 89.1 & 62.8 & 90.7 \\
      CLIP (ViT-L)~\cite{radford2021learning}    
        & 51.4 & 82.9 & 63.5 & 91.7 \\
      EVA-02-CLIP (ViT-L-336)~\cite{fang2023eva02}
        & 70.1 & 92.5 & 76.7 & 95.5 \\
      OpenCLIP (ViT-G/14)~\cite{NEURIPS2022_a1859deb}  
        & 76.0 & 95.1 & 76.7 & 96.3 \\
      OpenCLIP (ViT-BigG/14)~\cite{NEURIPS2022_a1859deb}
        & 81.9 & 96.1 & 82.0 & 97.6 \\

      \midrule
      Long-CLIP (ViT-L)~\cite{Zhang2024LongCLIPUT}
        & 86.1 & 96.2 & 82.7 & 96.4 \\
      TULIP (ViT-L)~\cite{najdenkoska2025tuliptokenlengthupgradedclip}
        & 91.1 & —    & 90.1 & —    \\
      \midrule
 \lexismallzs & 86.9 & 98.2 & 84.4 & 97.5 \\

\lexismallft{} & \textbf{97.1} & \textbf{99.9} & \textbf{96.7} & \textbf{100} \\

      \bottomrule
    \end{tabular}%
  }
  \label{table:urban1k_results}
  \vspace{-0.3cm}
\end{table}

\subsection{Image-text compositionality}
We evaluate our \name{} models on compositionality on the SugarCrepe \cite{NEURIPS2023_63461de0} and SugarCrepe++ \cite{NEURIPS2024_200661bf} benchmarks. As Tab.~\ref{table:sugarcrepe_results} shows, even without fine-tuning, \lexismallinline{} compares favorably against CLIP, outperforming the similarly sized \texttt{ViT-L} model. Our proposed finetuning process further improves the results by +7.2\% pts on average, surpassing even the much larger 2.5B (BigG/14) model and achieving state-of-the-art performance. We note that the largest gains are the Swap tasks—object up +5.3\% pts, attribute up +22.7\% pts-which directly probe ``bag-of-words'' shortcuts.\footnote{We provide the detailed results in the appendix.} Similar results can be observed on SugarCrepe++ in Tab. \ref{table:sugarcrepepp_results} 

\begin{table}[!ht]
\centering
\caption{Comparison with state-of-the-art on the SugarCrepe compositionality benchmark.}
\scriptsize
\setlength{\tabcolsep}{1pt}   
\renewcommand{\arraystretch}{1.05}
\resizebox{\columnwidth}{!}{%
\begin{tabular}{@{}l c c c c c@{}}
\toprule
\textbf{Method} 
  & \textbf{Params (B)} 
  & \textbf{Replace} 
  & \textbf{Swap} 
  & \textbf{Add} 
  & \textbf{Avg.} \\
\midrule
CLIP (ViT-B)~\cite{radford2021learning} & 0.15 & 80.1 & 62.7 & 73.0 & 71.9 \\
SigLIP ViT-B/16~\cite{zhai2023sigmoid}  & 0.23 & 84.1 & 65.7 & 86.4 & 78.7 \\
CLIP (ViT-L)~\cite{radford2021learning} & 0.43 & 79.5 & 61.3 & 74.9 & 71.9 \\
EVA-02-CLIP (ViT-L-336)~\cite{fang2023eva02} & 0.43 & 84.2 & 65.1 & 89.2 & 79.5 \\
BLIP (ViT-L)~\cite{li2022blip}          & 0.23 & 82.4 & 71.7 & 88.6 & 80.9 \\
BLIP2 (ViT-L)~\cite{li2023blip2}        & 1.17 & 85.7 & 63.8 & 89.9& 79.8 \\
OpenCLIP (ViT-G/14)~\cite{NEURIPS2022_a1859deb}   & 1.37 & 84.4 & 67.1 & 86.8 & 79.4 \\
OpenCLIP (ViT-BigG/14)~\cite{NEURIPS2022_a1859deb}& 2.54 & 86.5 & 68.9 & 88.4 & 81.3 \\
\midrule
NegCLIP~\cite{yuksekgonul2023visionlanguagemodelsbehavelike}   & 0.15 & 85.0 & \textbf{75.3} & 85.8 & 82.0 \\
\midrule
\lexismallzs & 0.3  & 85.3 & 61.6 & 85.4 & 77.4 \\
\lexismallft{} & 0.3 & \textbf{86.8} & \textbf{75.6} & \textbf{91.3} & \textbf{84.6} \\

\bottomrule
\end{tabular}%
}
\label{table:sugarcrepe_results}
\vspace{-0.4cm}
\end{table}

\begin{table*}[!ht]
\centering
\caption{Comparison with state-of-the-art on the SugarCrepe++ compositionality benchmark.}
\scriptsize
\setlength{\tabcolsep}{1.5pt}   
\renewcommand{\arraystretch}{1.1}
\resizebox{0.9\textwidth}{!}{%
\begin{tabular}{@{}lccccccccccccc@{}}
\toprule
\textbf{Method} & \textbf{Params} & \multicolumn{2}{c}{\textbf{Swap Object}} & \multicolumn{2}{c}{\textbf{Swap Attribute}} & \multicolumn{2}{c}{\textbf{Replace Object}} & \multicolumn{2}{c}{\textbf{Replace Attribute}} & \multicolumn{2}{c}{\textbf{Replace Relation}} & \textbf{Avg.} & \textbf{Avg.} \\
& \textbf{(B)} & ITT & TOT & ITT & TOT & ITT & TOT & ITT & TOT & ITT & TOT & ITT & TOT \\
\midrule
Human & -- & 100.00 & 96.7 & 96.7 & 93.3 & 100.00 & 97.0 & 100.00 & 98.3 & 100.00 & 96.7 & 99.3 & 96.4 \\
\midrule
CLIP (ViT-B)~\cite{radford2021learning} & 0.15 & 45.2 & 19.7 & 45.2 & 33.0 & 86.8 & 83.7 & 65.6 & 59.1 & 56.3 & 38.6 & 59.8 & 46.8 \\
SigLIP ViT-B/16~\cite{zhai2023sigmoid} & 0.23 & 39.5 & 23.0 & 56.1 & 46.4 & 91.3 & 79.2 & 75.2 & 64.0 & 54.8 & 45.0 & 63.4 & 51.5 \\
CLIP (ViT-L)~\cite{radford2021learning} & 0.43 & 46.0 & 14.5 & 44.5 & 28.7 & 92.0 & 81.3 & 68.8 & 56.3 & 53.4 & 39.1 & 60.6 & 44.0 \\
EVA-02-CLIP (ViT-L-336)~\cite{fang2023eva02} & 0.43 & 44.1 & 19.2 & 47.3 & 34.4 & 94.2 & 91.6 & 74.5 & 69.5 & 59.8 & 48.9 & 64.0 & 52.7 \\
BLIP (ViT-L)~\cite{li2022blip} & 0.23 & 46.8 & 29.8 & 60.1 & 52.5 & 92.6 & 89.1 & 71.7 & 75.0 & 56.8 & 57.7 & 65.6 & 60.8 \\
BLIP2 (ViT-L)~\cite{li2023blip2} & 1.17 & 37.9 & 39.5 & 51.9 & 55.4 & \textbf{94.8} & 96.9 & 73.2 & 86.5 & 65.1 & 69.6 & 64.6 & 69.6 \\
OpenCLIP (ViT-G/14)~\cite{NEURIPS2022_a1859deb} & 1.37 & 40.7 & 27.4 & 54.2 & 49.6 & 93.1 & 89.4 & 72.5 & 73.1 & 57.6 & 51.4 & 63.6 & 58.2 \\
OpenCLIP (ViT-BigG/14)~\cite{NEURIPS2022_a1859deb} & 2.54 & 48.8 & 28.2 & 57.7 & 52.4 & 94.2 & 90.5 & 76.4 & 72.6 & 59.4 & 53.6 & 67.3 & 59.5 \\
\midrule
NegCLIP~\cite{yuksekgonul2023visionlanguagemodelsbehavelike} & 0.15 & \textbf{55.3} & 34.7 & 58.0 & 56.5 & 89.5 & 94.5 & 69.4 & 76.3 & 52.3 & 51.6 & 64.9 & 62.7 \\
CLIP-SVLC~\cite{Doveh2022TeachingSV} & 0.15 & 43.0 & 18.9 & 48.4 & 34.6 & 80.9 & 91.6 & 57.0 & 66.9 & 47.3 & 51.3 & 55.3 & 52.7 \\
BLIP-SGVL~\cite{herzig2023incorporatingstructuredrepresentationspretrained} & 0.15 & 13.2 & --  & 38.8 & --  & 53.8 & --  & 34.4 & --  & 30.7 & --  & 34.2 & --    \\
\midrule
\lexismallzs & 0.3 & 48.2 & 20.8 & 43.8 & 28.1 & 91.2 & 95.6 & 75.5 & 85.8 & \textbf{72.4} & \textbf{77.2} & 66.2 & 61.5 \\
\lexismallft{} & 0.3 & \textbf{53.9} & \textbf{43.3} & \textbf{68.3} & \textbf{68.3} & 93.8 & \textbf{97.3} & \textbf{77.2} & \textbf{88.7} & 65.9 & 71.6 & \textbf{71.8} & \textbf{73.8} \\
\bottomrule
\end{tabular}%
}
\label{table:sugarcrepepp_results}
\end{table*}

\section{Ablation studies and analysis}
\label{sec:ablations}
\subsection{Bridging The Modality Gap}

The modality gap~\cite{liang2022mind} is one of the primary factors contributing to the poor compositionality of contrastive vision-language models and can also adversely affect model fairness. We assess below, on the Flickr test set, its evolution from the initial zero-shot configuration through to the post-finetuning stage by analyzing the distribution of pairwise cosine similarities and the inter-modality distance between image and text representations.

\begin{figure}[ht]
  \centering
  \begin{subfigure}[b]{0.45\linewidth}
    \includegraphics[width=\linewidth]{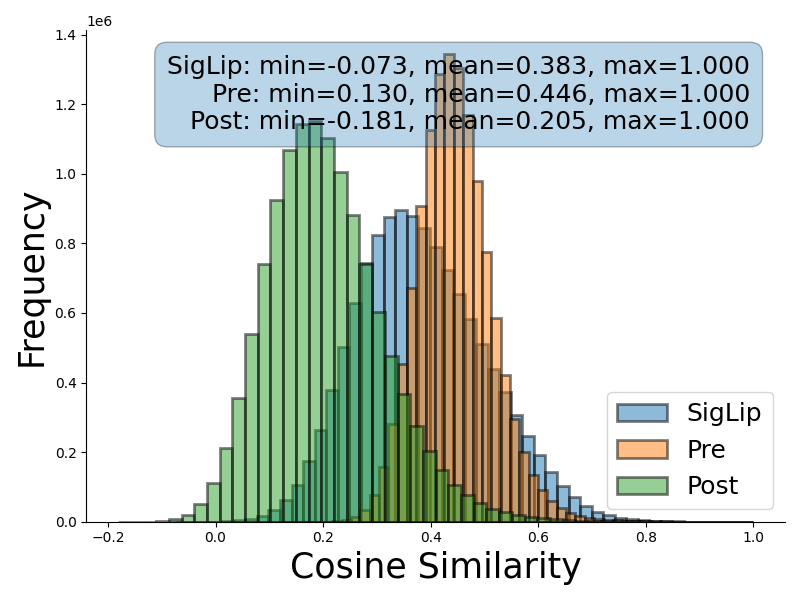}
    \caption{Text embeddings}
    \label{fig:text-cosine}
  \end{subfigure}
~
  \begin{subfigure}[b]{0.45\linewidth}
    \includegraphics[width=\linewidth]{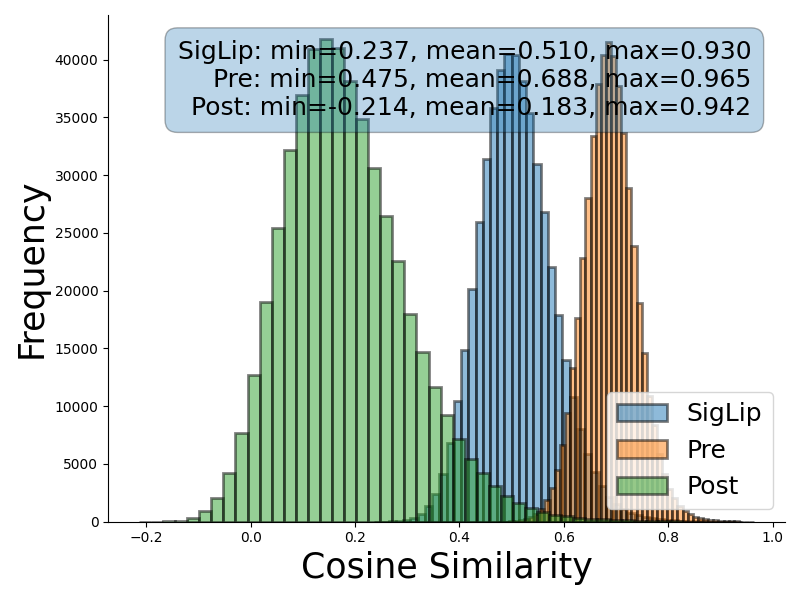}
    \caption{Image embeddings}
    \label{fig:image-cosine}
  \end{subfigure}
  \vspace{-0.2cm}
  \caption{
    Distributions of pairwise cosine similarities for three embedding sets: 
    \textbf{SigLip} (blue), \emph{\lexismallzs}(orange), and \emph{\lexismallft} (green). 
  }
  \label{fig:cosine-distributions}
  \vspace{-0.5cm}
\end{figure}

\noindent \textbf{Pairwise Cosine Similarity Distributions:} Fig.~\ref{fig:cosine-distributions} presents the pairwise cosine-similarity distributions for three models: SigLip, \lexismallinline before finetuning, and  \lexismallinline after fine-tuning.  For both modalities, the pre-distributions are tightly concentrated at high similarity, indicating a partially ``collapsed'' space where unrelated pairs remain largely aligned. After fine-tuning, the distributions shift toward lower similarities and broaden, demonstrating that the model has learned to \emph{de-collapse} its representation space, pushing unrelated instances farther apart.  This increased spread is especially pronounced in the image modality, where the Post histogram has both peaks at lower cosine values and extends over a wider band, suggesting that visual features benefit strongly from fine-tuning in terms of discriminative power.

\begin{figure*}[!ht]
\vspace{-0.0cm}
  \centering
  \begin{subfigure}[b]{0.3\linewidth}
    \includegraphics[width=\linewidth]{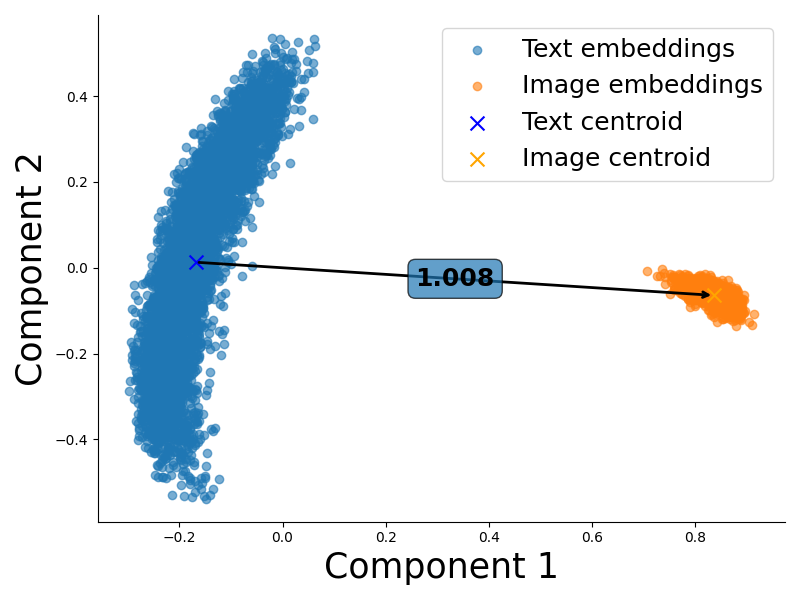}
    \caption{SigLip (zero-shot)}
    \label{fig:gap-siglip}
  \end{subfigure}\hfill
  \begin{subfigure}[b]{0.3\linewidth}
    \includegraphics[width=\linewidth]{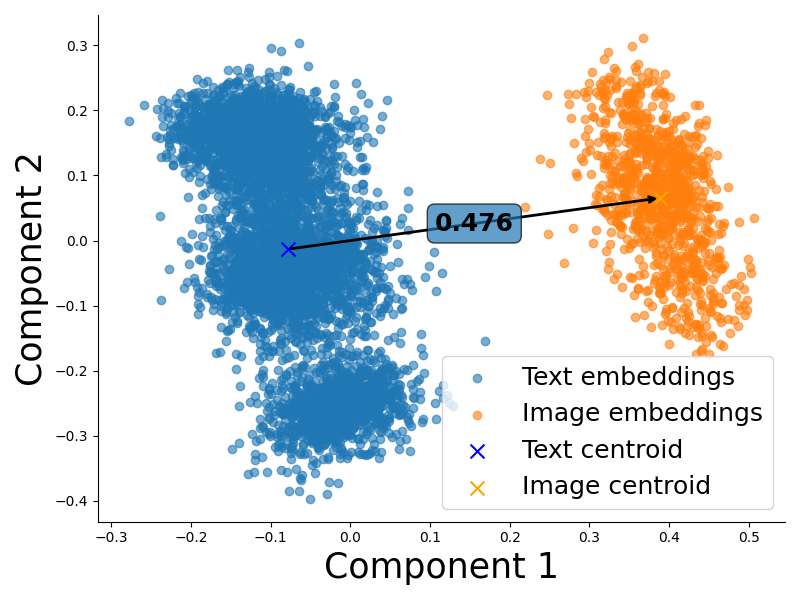}
    \caption{Our model, pre–finetuning}
    \label{fig:gap-pre}
  \end{subfigure}\hfill
  \begin{subfigure}[b]{0.3\linewidth}
    \includegraphics[width=\linewidth]{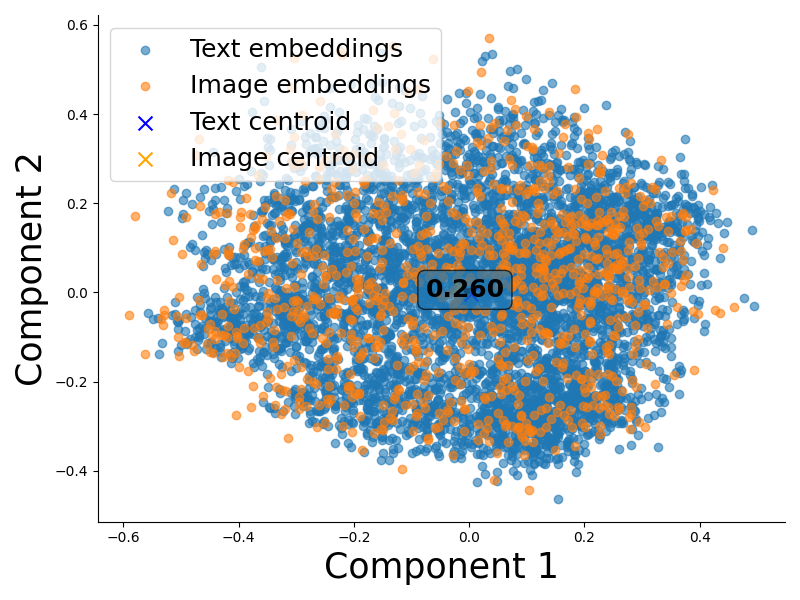}
    \caption{Our model, post–finetuning}
    \label{fig:gap-post}
  \end{subfigure}
  \vspace{-0.3cm}
  \caption{%
    2D (PCA) projections of text vs.\ image embeddings, with centroids (×) and the modality‐gap arrow annotated by its Euclidean length.  
    (a) SigLip exhibits a large gap of $\approx1.008$,  
    (b) our model before fine‐tuning has a gap of $\approx0.476$,  
    (c) after fine‐tuning the gap shrinks to $\approx0.260$, indicating improved alignment between modalities.%
  }
  \label{fig:modality-gap-comparison}
  \vspace{-0.2cm}
\end{figure*}

\noindent \textbf{Modality Gap Comparison:} Fig. \ref{fig:modality-gap-comparison} demonstrates that fine-tuning narrows the distance between text and image embeddings in the shared space, as shown by their projection into two dimensions via PCA.
In the SigLip baseline (Fig.~\ref{fig:gap-siglip}), the centroids of text and image representations are separated by $\sim1.008$, reflecting a substantial modality gap. With our method, even prior to target-task finetuning (Fig.~\ref{fig:gap-pre}), this gap is reduced to $0.476$, a significant improvement attributable to the unimodal architecture. Crucially, after finetuning  (Fig.~\ref{fig:gap-post}), the centroid distance further decreases to $0.260$. This final gap is nearly half that of our model before finetuning and roughly a quarter of the original SigLip gap. This progressive narrowing demonstrates two key points: (1) our initial zero-shot alignment significantly improves upon the SigLip, and (2) the subsequent finetuning further tightens modality alignment, enhancing the cross-modal retrieval performance.
\label{sec:modality-gap}

\begin{table*}[!ht]
\centering
\caption{Zero‐shot retrieval accuracy and compositional understanding under two different ablations.}
\scriptsize
\setlength{\tabcolsep}{1pt}
\vspace{-0.3cm}
\renewcommand{\arraystretch}{1.1}
\begin{subtable}[t]{0.7\textwidth}
  \centering
  \caption{With vs.\ without object‐based descriptions}
  \vspace{-0.2cm}
  \resizebox{\textwidth}{!}{%
\begin{tabular}{@{}lc *{3}{cc} *{5}{c}@{}}
\toprule
\textbf{Model} & \textbf{Obj.} 
& \multicolumn{2}{c}{\textbf{Flickr30K}}
& \multicolumn{2}{c}{\textbf{COCO}}
& \multicolumn{2}{c}{\textbf{Urban1k}}
& \multicolumn{5}{c}{\textbf{SugarCrepe++ (ITT)}} \\
\cmidrule(lr){3-4}\cmidrule(lr){5-6}\cmidrule(lr){7-8}\cmidrule(lr){9-13} 
& \textbf{Desc.} & R@1 & R@10 & R@1 & R@10 & R@1 & R@10
& Swap Obj & Swap Attr & Repl Obj & Repl Attr & Repl Rel \\
\midrule
\lexismallzs  & $\checkmark$ 
& 69.5 & 94.2 & 41.7 & 76.6 & 86.9 & 98.2
& 48.2 & 43.8 & 91.2 & 75.5 & 72.4 \\
\lexilargezs  & $\checkmark$ & 74.4 & 95.1 & 46.4 & 80.2 & 91.8 & 99.2 & 49.8 & 57.2 & 91.5 & 77.9 & 77.7 \\
\midrule 
\lexismallzs & 
& 66.4 & 92.8 & 38.7 & 73.8 & 83.7 & 96.5
& 47.3 & 42.5 & 88.6 & 76.6 & 71.9 \\
\lexilargezs & 
& 72.1 & 94.5 & 43.4 & 78.0 & 89.2 & 98.7
& 49.0 & 56.2 & 89.0 & 78.9 & 77.2 \\
\bottomrule
\end{tabular}
  }
  \label{tab:ablation_desc}
\end{subtable}%
\hfill
\begin{subtable}[t]{0.7\textwidth}
  \centering
  \caption{Max sequence‐length ablation}
  \vspace{-0.2cm}
  \resizebox{\textwidth}{!}{%
    \begin{tabular}{@{}l c *{3}{cc} *{5}{c}@{}}
    \toprule
    \textbf{Model} & \textbf{Max Seq}
      & \multicolumn{2}{c}{\textbf{Flickr30K}}
      & \multicolumn{2}{c}{\textbf{COCO}}
      & \multicolumn{2}{c}{\textbf{Urban1k}}
      & \multicolumn{5}{c}{\textbf{SugarCrepe++ (ITT)}} \\
    \cmidrule(lr){3-4}\cmidrule(lr){5-6}\cmidrule(lr){7-8}\cmidrule(lr){9-13}
      &  \textbf{Len} & R@1 & R@10 & R@1 & R@10 & R@1 & R@10
      & Swap Obj & Swap Attr & Repl Obj & Repl Attr & Repl Rel \\
    \midrule

\lexismallzs& 256
      & 69.5 & 94.2 & 41.7 & 76.6 & 86.9 & 98.2
      & 48.2 & 43.8 & 91.2 & 75.5 & 72.4 \\

      \textbf{\name{} (7B)-ZS}& 256
      & 74.4 & 95.1 & 46.4 & 80.2 & 91.8 & 99.2
      & 49.8 & 57.2 & 91.5 & 77.9 & 77.7 \\
    \midrule
      \lexismallzs& 512
      & 69.1 & 94.0 & 41.1 & 76.3 & 86.9 & 98.2
      & 47.8 & 44.4 & 91.3 & 75.0 & 71.2 \\
    \textbf{\name{} (7B)-ZS}    & 1024
      & 74.2 & 95.0 & 46.1 & 80.2 & 92.7 & 99.3
      & 49.4 & 57.8 & 91.1 & 78.9 & 77.8 \\
    \bottomrule
    \end{tabular}%
  }
  \label{tab:ablation_seq}
\end{subtable}
\label{tab:merged_ablation}
\vspace{-0.5cm}
\end{table*}

\subsection{Impact of the proposed components}
To better understand the key components of our data-to-text pipeline, we ablate the impact of a) object-based descriptions and b) length of the image description. Zero-shot image retrieval (on Flickr30k, COCO, Urban1K) and compositional understanding (on SugarCreppe++) are evaluated for the 0.3B \lexismallzs~ and a much bigger 7B parameter decoder-only model based on BGE-en-ICL-7B~\cite{li2024makingtextembeddersfewshot}, denoted as \lexilargezs{}. Based on our experiments, we draw the following conclusions:

\begin{table}[!h]
\centering
\caption{Zero-shot retrieval on Flickr30K: Impact of the VLLM captioner.}
\vspace{-0.4cm}
\scriptsize
\setlength{\tabcolsep}{2pt}
\renewcommand{\arraystretch}{1.1}

\begin{subtable}{\columnwidth}
\centering
\caption{Effect of VLLM architecture.}
\resizebox{\columnwidth}{!}{%
  \begin{tabular}{@{}l c c c c c@{}}
    \toprule
    \textbf{Method} & \textbf{VLLM} & \multicolumn{2}{c}{\textbf{Image retrieval}} & \multicolumn{2}{c}{\textbf{Text retrieval}} \\
    \cmidrule(lr){3-4}\cmidrule(lr){5-6}
    & & R@1 & R@10 & R@1 & R@10 \\
    \midrule
    \lexismallzs & InternVL2.5-8B-MPO & 69.5 & \textbf{94.2} & 75.9 & \textbf{97.4} \\
    \lexismallzs & Qwen2.5-VL-7B      & 65.8 & 92.6 & 67.8 & 96.0 \\
    \lexismallzs & MiniCPM-V-2\_6-8B  & 67.2 & 92.9 & \textbf{75.3} & 96.6 \\
    \bottomrule
  \end{tabular}
}
\label{subtab:vllm_type}
\end{subtable}

\vspace{0.2cm}

\begin{subtable}{\columnwidth}
\centering
\caption{Effect of VLLM size.}
\resizebox{\columnwidth}{!}{%
  \begin{tabular}{@{}l c c c c c c@{}}
    \toprule
    \textbf{Method} & \textbf{VLLM} & \textbf{Size (B)} & \multicolumn{2}{c}{\textbf{Image retrieval}} & \multicolumn{2}{c}{\textbf{Text retrieval}} \\
    \cmidrule(lr){4-5}\cmidrule(lr){6-7}
    & & & R@1 & R@10 & R@1 & R@10 \\
    \midrule
    \lexismallzs & InternVL2.5-MPO & 1  & 69.4 & 94.0 & 75.8 & 97.1 \\
    \lexismallzs & InternVL2.5-MPO & 2  & 68.2 & 94.0 & 74.3 & 97.2 \\
    \lexismallzs & InternVL2.5-MPO & 4  & \textbf{70.9} & 93.9 & \textbf{78.6} & 96.9 \\
    \lexismallzs & InternVL2.5-MPO & 8  & 69.5 & 94.2 & 75.9 & 97.4 \\
    \lexismallzs & InternVL3       & 9  & 70.7 & \textbf{94.6} & 77.9 & \textbf{97.7} \\
    \lexismallzs & InternVL3       & 14 & 70.3 & 94.3 & 74.2 & 97.2 \\
    \bottomrule
  \end{tabular}
}
\label{subtab:vllm_size2}
\end{subtable}

\vspace{0.2cm}

\begin{subtable}{\columnwidth}
\centering
\caption{Effect of cross-VLLM inference (different captioners at training vs.~inference).}
\resizebox{\columnwidth}{!}{%
  \begin{tabular}{@{}l c c c c c c@{}}
    \toprule
    \textbf{Method} & \textbf{VLLM} & \textbf{Size (B)} & \multicolumn{2}{c}{\textbf{Image retrieval}} & \multicolumn{2}{c}{\textbf{Text retrieval}} \\
    \cmidrule(lr){4-5}\cmidrule(lr){6-7}
    & & & R@1 & R@10 & R@1 & R@10 \\
    \midrule
    \lexismallft & InternVL2.5-MPO & 1  & 78.7 & 97.1 & 90.8 & 99.5 \\
    \lexismallft & InternVL2.5-MPO & 2  & 77.9 & 96.7 & \textbf{92.0} & 99.6 \\
    \lexismallft & InternVL2.5-MPO & 4  & \textbf{79.3} & \textbf{97.6} & \textbf{92.0} & 99.4 \\
    \lexismallft & InternVL2.5-MPO & 8  & 79.2 & 97.4 & 91.6 & \textbf{99.7} \\
    \lexismallft & InternVL3       & 9  & \textbf{79.5} & 97.2 & 91.9 & \textbf{99.9} \\
    \lexismallft & InternVL3       & 14 & 78.7 & 97.0 & \textbf{92.0} & 99.5 \\
    \bottomrule
  \end{tabular}
}
\label{subtab:cross_vllm_type}
\end{subtable}

\label{tab:all_ablation_flickr}
\end{table}

\noindent \textbf{Object-based descriptions lead to improved accuracy}: As the results from Tab.~\ref{tab:ablation_desc} show, the addition of object-based descriptions consistently enhances performance across all evaluated datasets and tasks, underscoring the importance of dense object-attribute coverage. 

\noindent \textbf{Longer descriptions do not bring improvements}: in Tab.~\ref{tab:ablation_seq} we report results for a 300M and 7B sized model on two different sequence lengths, 256 tokens and 512/1024 for the 300M/7B model. For the smaller 300M model, increasing the sequence length from 256 to its maximum of 512 tokens yields minimal gains. Similarly, the larger 7B model remains largely stable with only minor gains on Urban1k. In general, 256 tokens suffice, and further increases do not demonstrate accuracy improvements.

\subsection{Captioner choices}
In Tab.~\ref{tab:all_ablation_flickr}, we ablate the impact of the captioner in zero-shot retrieval of Flickr30k. 

\noindent \textbf{VLLM architecture}: In Tab.~\ref{subtab:vllm_type} we evaluate how the choice of the image-to-text VLLM convertor impacts the downstream performance. We evaluate three similarly sized state-of-the-art VLLM,  Qwen2.5-VL-7b \cite{bai2025qwen25vltechnicalreport}, MiniCPM-V-2-6-8b \cite{yao2024minicpm} and InternVL-2.5-8b-MPO. We found InternVL to perform better. This highlights that better VLLM results in higher performance and that careful consideration should be made when choosing the VLLM.

\noindent \textbf{Effect of Size of the Image-to-Text VLLM Convertor}: In Tab.~\ref{subtab:vllm_size2} we ablate models ranging from 1B to 14B parameters as captioners. We note that our approach is robust to the size of the image captioner. Also, we observe that the 1B model performs nearly as well as the 8B and 14B models, showcasing efficient performance even with smaller models.

\noindent \textbf{Captioner-Agnostic Inference}: In Tab.~\ref{subtab:cross_vllm_type}, we investigate the transferability of \name~(0.3B)-FT across different caption models. Our model, fine-tuned once on captions from the 8B InternVL model, maintains its performance even when switching to different image description sources (e.g., using a 1B model instead of 8B) without further fine-tuning. This indicates that we can fine-tune the model only once and then change the image captioner freely.

\subsection{Information loss when converting Images to Text}

Departing from raw pixels to text descriptions, even if they are detailed, inherently risks losing subtle visual information. To estimate how well our generated image descriptions capture visual content, we measure the overlap with the ground-truth object classes in MS-COCO (80 classes). With exact class-name matching, we get $69\%$  match score between objects in our descriptions and those in ground truth annotations.  However, we note that possibly there is a higher overlap, as our estimation does not account for synonyms or paraphrasing.

\section{Conclusions}

We introduce a text-to-text paradigm for training a vision-free single-encoder CLIP model, challenging the conventional two-tower paradigm. Our framework uses VLLMs to generate structured image descriptions and omits images during training. This reduces the modality gap and improves compositional generalisation while achieving better performance on short caption queries.  Unlike traditional two-tower architectures, our \name is able to effectively model the full spectrum of query distributions, from brief user-centric queries to long, paragraph-level descriptions -  all using the same single encoder. To further assess compositional generalisation, we release subFlickr and subCOCO, two curated benchmarks with diverse compositional queries made of short captions. Finally, we show that models with 0.3B parameters can match or even surpass traditional multimodal architectures, achieving SOTA results across multiple compositionality benchmarks and retrieval tasks.

\section{Limitations}

The limitations of this work are mostly related to its strong dependence on VLLMs. As we depart from raw pixels to text-based image descriptions generated by VLLMs, certain visual details will probably be lost. In particular, descriptions of crowded scenes or those containing many small objects are likely to omit a significant amount of information. Moreover, since VLLMs are often biased or have hallucinations, generated descriptions often inherit them. Such errors can potentially propagate into the retrieval process. As future work, to mitigate these issues, we will try to adopt filtering methods, ensemble captioners, or even some kind of human-in-the-loop verification. Then, another limitation of using VLLMs as an image captioner is the extra computational overhead introduced. We calculate that for an A100 GPU, an unoptimized implementation requires approximately 0.2 seconds per image. This cost can be significantly reduced through optimized implementations (e.g., the \texttt{vllm} project) and techniques like quantization. In our work, this step is performed once, offline, ahead of evaluation, while retrieval itself remains efficient. On an A100 GPU, our method requires only 1.6 ms per image, compared to 7.4 ms for OpenCLIP-2.54B. Finally, we note that similar retrieval performance can be obtained with smaller generators (e.g., a 2B InternVL), which can lower the preprocessing cost.

\section{Broader Impact}

We also reflect on the broader impact and ethics of our work. Given that the main body of our retrieval pipeline is based on text rather than images, we consider that \name allows for a more interpretable and transparent retrieval. Additionally, \name is a more inclusive approach for users with visual impairments. However, we acknowledge that the heavy reliance of our work,  in VLLMs to generate image descriptions, is accompanied by inherited biases that may reinforce stereotypes or amplify unfair associations. 

 \section{Acknowledgments}

This work was partially funded by UK Research and Innovation (UKRI) under the UK government’s Horizon Europe funding guarantee (grant No. 10099264) and by the European Union (under EC Horizon Europe grant agreement No. 101135800 (RAIDO)). 
\bibliography{custom}

\newpage
\appendix

\section{Technical Appendices and Supplementary Material}

\subsection{subFlickr and subCoco benchmarks}

\begin{figure}[ht]
    \centering
    \begin{subfigure}{\linewidth}
        \centering
        \includegraphics[width=\linewidth]{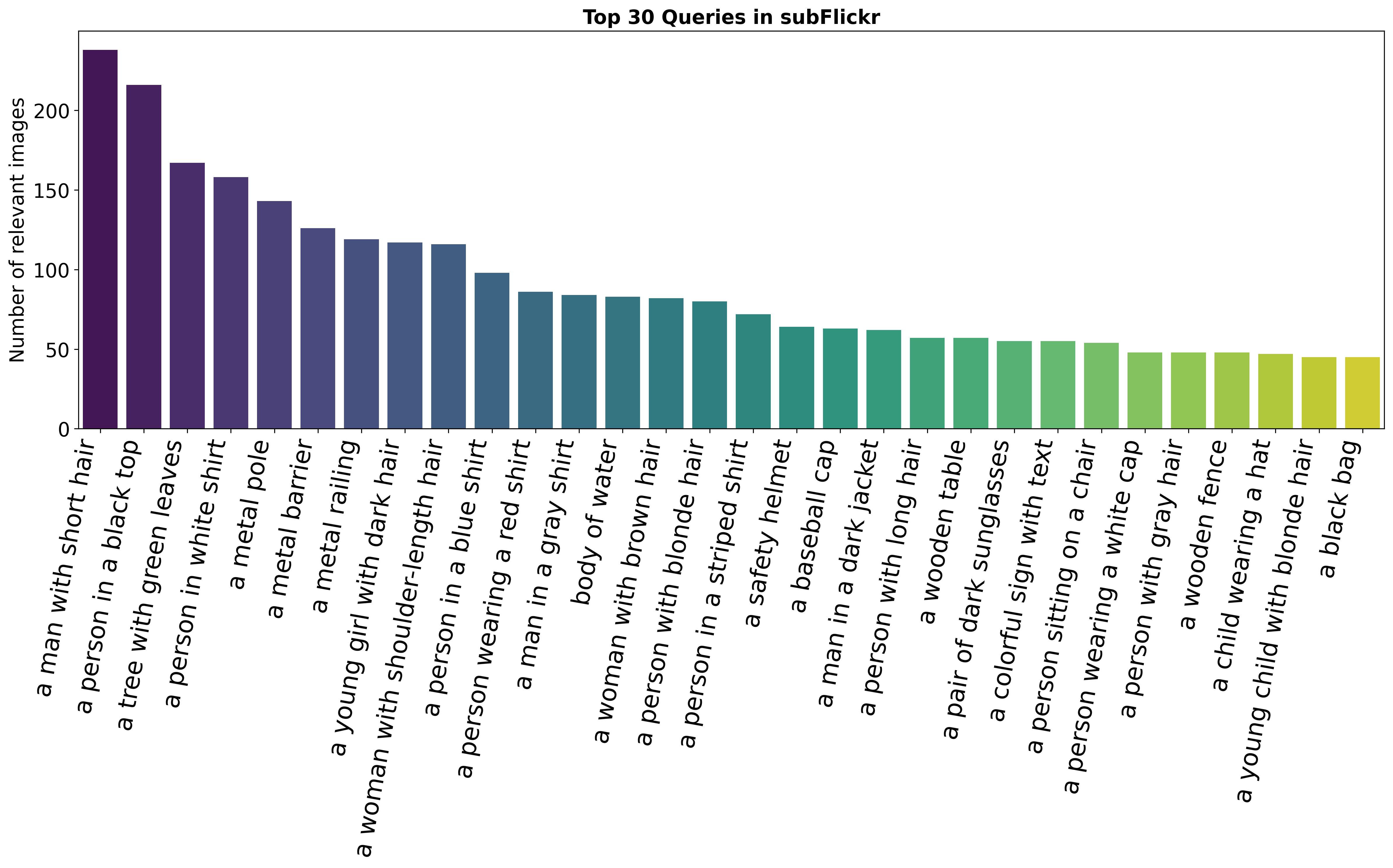}
        \caption{Top 30 most frequent queries in the \textit{subFlickr} dataset.}
        \label{fig:subflickr_query_hist}
        \vspace{0.5em}
    \end{subfigure}

    \begin{subfigure}{\linewidth}
        \centering
        \includegraphics[width=\linewidth]{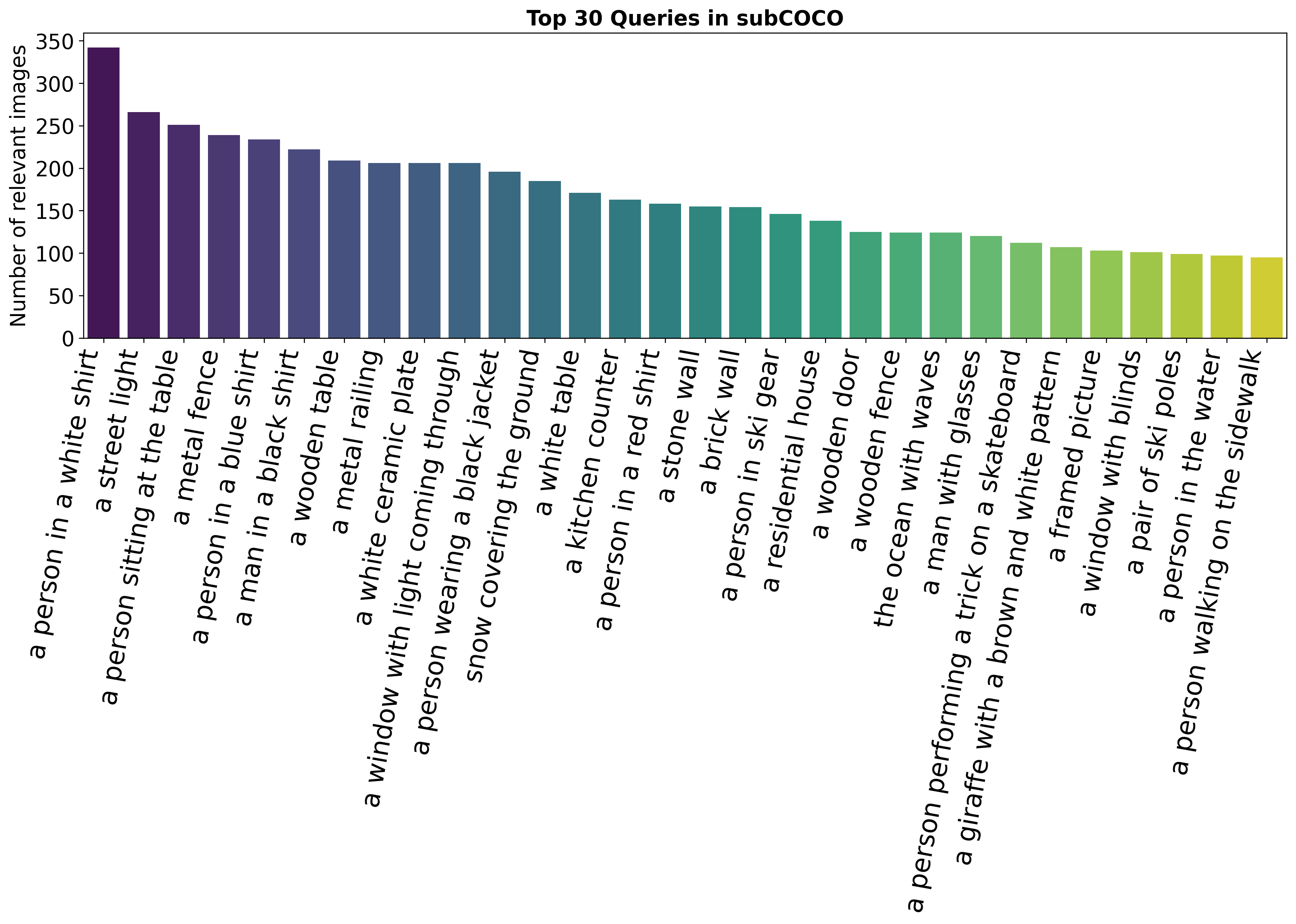}
        \caption{Top 30 most frequent queries in the \textit{subCOCO} dataset.}
        \label{fig:subcoco_query_hist}
    \end{subfigure}

    \caption{Query frequency distributions in the \textit{subFlickr} (a) and \textit{subCOCO} (b) datasets. The y-axis indicates how many images are relevant to each query.Looks better zoomed in.}
    \label{fig:query_freq_combined}
\end{figure}

Fig.~\ref {fig:query_freq_combined} shows the distribution of the top 30 most frequent queries in the \textit{subFlickr} and \textit{subCOCO}
datasets. Both datasets have long-tail distribution with \textit{subCOCO} showing a slightly sharper drop in frequency. 

\subsection{Representing images using text}

\begin{figure*}[ht]
  \centering
  \includegraphics[
    width=\textwidth
  ]{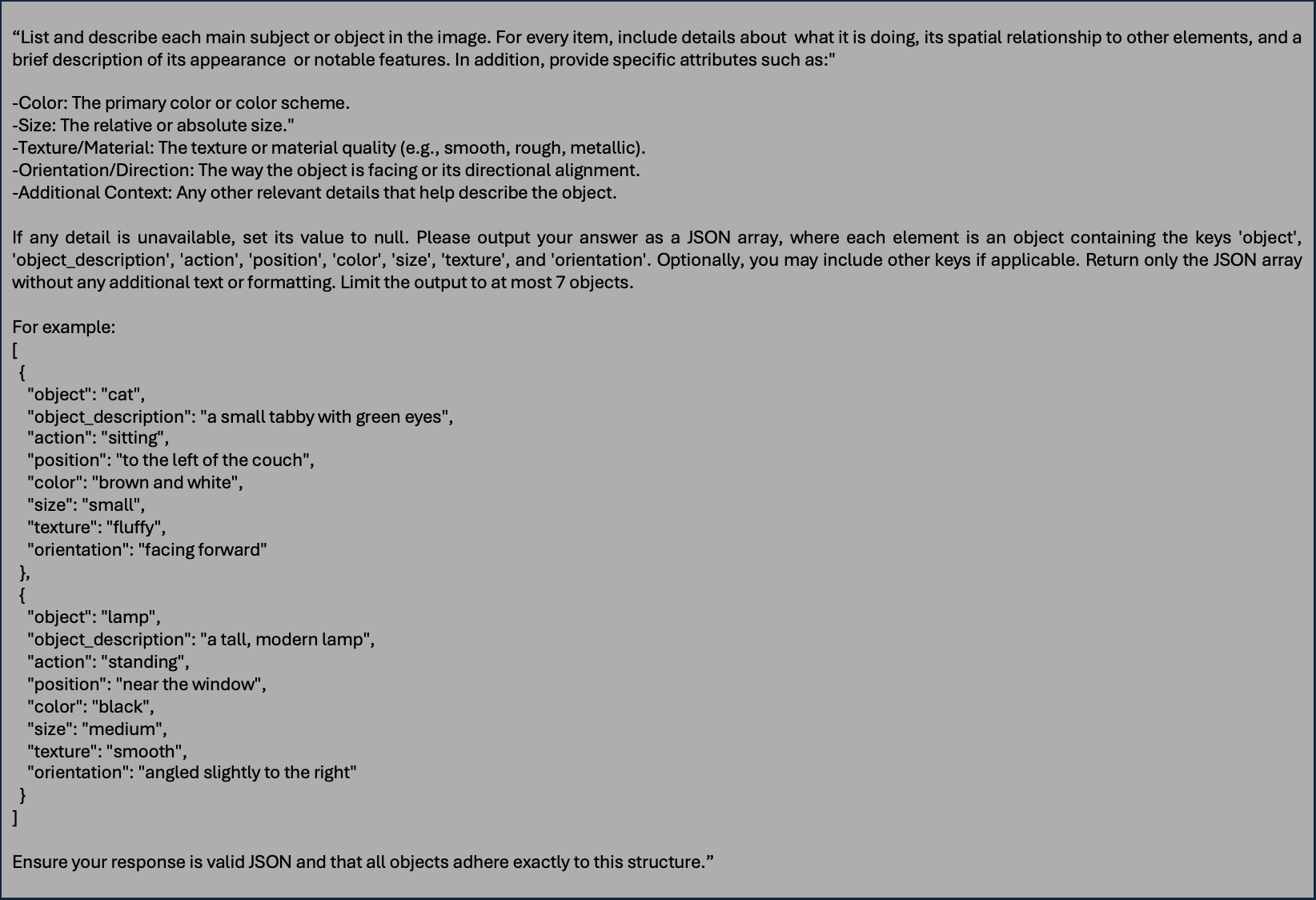}
  \caption{The prompt that was given to the model for extracting the object annotations.}
  \label{fig:prompt_for_object_annots}
  \vspace{-0.4cm}
\end{figure*}

To convert images into rich, structured text, we employ the OpenGVLab/InternVL2\_5-8B-MPO model\footnote{\url{https://huggingface.co/OpenGVLab/InternVL2\_5-8B-MPO}}. We extract two complementary views of each image:

\begin{itemize}
\item A \emph{detailed scene description}, obtained by prompting the model with:
\begin{quote}
\texttt{Please describe the image in detail.}
\end{quote}
\item \emph{Object annotations}, generated using the prompt shown in Fig.~\ref{fig:prompt_for_object_annots}.
\end{itemize}

\begin{figure*}[!htbp]
  \centering
  \includegraphics[
    width=\textwidth,
    keepaspectratio,
    trim=10 10 10 10,
    clip
  ]{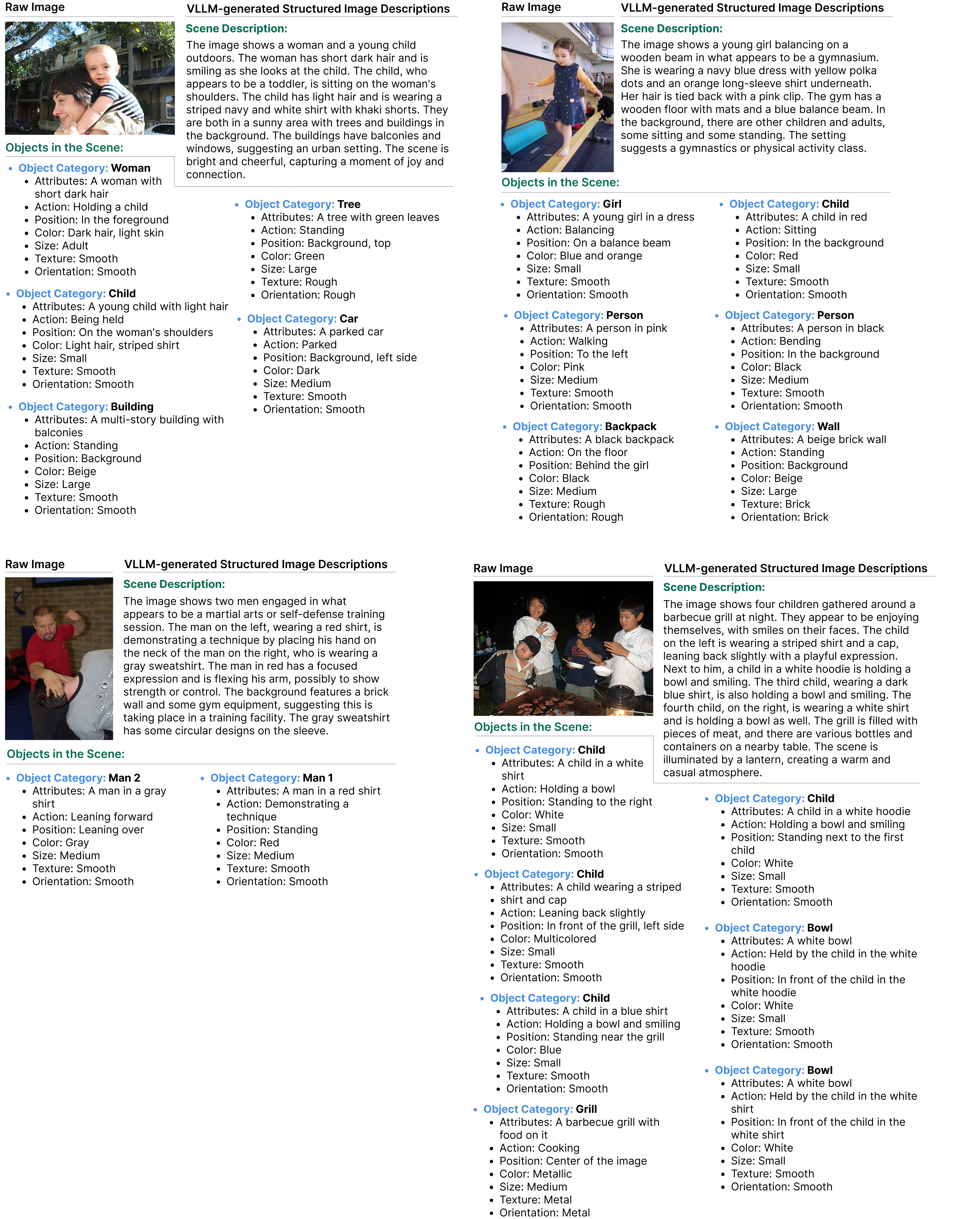}
  \caption{Examples of our image representation in text: on the left, the raw image; on the right, the corresponding structured description generated by our model.}
  \label{fig:image_rep_example_appendix}
  \vspace{-0.4cm}
\end{figure*}

Fig.~\ref{fig:image_rep_example_appendix} presents several examples of this text‐based representation—on the left, the raw image; on the right, our model’s concatenated scene summary and per‐object attribute list.

\subsection{Generation of compositional captions}
As stated in ~\ref{ssec:implementation},  we train our pipeline in two stages. Given that OpenImages is not paired with longer or compositional captions, we synthetically generate them. For concise captions, we run BLIP-2. For smaller compositional captions, we ask InternVL-2.5-8B-MPO to generate a pool of  $6$ compositional captions for each image based on its generated image description in order to accelerate the data preparation process. Below, we provide the prompt we have used.

\begin{quote}
You are given a scene caption and a list of structured object descriptions extracted from an image. 
Your task is to generate 6 short compositional search queries (2 to 4 words) that someone might use to find this image. 
Each query should refer to an object, attribute, or visual element described in the input. 

Use combinations like:
- object + color (e.g., ``red bird'')
- object + position (e.g., ``bird on branch'')
- object + action (e.g., ``bird flying'')
- or noun phrases (e.g., ``bird cage'')

Return only a raw JSON list of 6 strings like:
[``black bird'', ``bird feeder'', ``bird in cage'', ``wooden birdhouse'', ``birds perched'', ``red flowers'']

Do not include markdown or code formatting such as triple backticks or \texttt{json} labels.
\end{quote}

\subsection{Training details}
We train our model in two stages. For both stages, the model is trained for three epochs using the FlagEmbedding library~\cite{bge_embedding}, with a cosine annealed learning rate schedule and a warm-up phase 5\%.  The model is trained on two A100 GPUs with an effective batch size of 2,048, a peak learning rate of $1\times 10^{-4}$ and a weight decay of 0.1. Training employs mixed precision (FP16), gradient checkpoint, and DeepSpeed. 

During the first stage, the training is done only using BLIP-2 captions using a symmetric contrastive loss. In the second stage, we fine-tune the model using a mixture of BLIP-2 and compositional captions, where each batch of 2,048 samples consists of 200 concise BLIP-2 captions and the remainder compositional ones. In this stag,e the loss is only text-to-image. The final model is obtained by averaging the checkpoints from the two stages, weighted 0.4 for the first and 0.6 for the second stage.

\section{Additional ablation studies}
We extend our ablation studies of Sect.~\ref{sec:ablations} to investigate two aspects. First, we assess the effect of the maximum sequence length on downstream retrieval in Tab.~\ref{tab:ablate_max_len_for_encoding}.
We note a clear performance drop for descriptions shorter than 128 and diminishing returns above 256.
Second, in Tab.~\ref{tab:ablate_size_of_vllm_7b_model_inference} we check the zero-shot retrieval performance on Flickr30K by varying the size of the VLLM used to generate image and object-level descriptions, while keeping the 7B model as the retriever.

\begin{table}[ht]
\centering
\caption{Zero‐shot retrieval on Flickr30K, ablating the max seq. len used for encoding the image and object descriptions.}
\vspace{-0.4cm}
\scriptsize
\setlength{\tabcolsep}{2pt}
\renewcommand{\arraystretch}{1.1}
\resizebox{\columnwidth}{!}{%
  \begin{tabular}{@{}l c c c c c@{}}
    \toprule
    \textbf{Method} & \textbf{Max Seq Len} & \multicolumn{2}{c}{\textbf{Image retrieval}} & \multicolumn{2}{c}{\textbf{Text retrieval}} \\
    \cmidrule(lr){3-4}\cmidrule(lr){5-6}
    & & R@1 & R@10 & R@1 & R@10 \\
    \midrule
\lexismallzs & 64  & 63.3 & 90.1 & 71.5 & 94.6 \\
\lexismallzs & 128 & 67.5 & 93.2 & 73.5 & 95.8 \\
\lexismallzs & 256 & \textbf{69.5} & \textbf{94.0} & \textbf{75.9} & \textbf{97.4} \\
\lexismallzs & 512 & 69.1 & 94.0 & 74.8 & 97.1 \\
    \bottomrule
  \end{tabular}%
}
\label{tab:ablate_max_len_for_encoding}
\end{table}

\begin{table}[ht]
\centering
\caption{Zero‐shot retrieval on Flickr30K, ablating the size of the VLLM used for extracting the image and object descriptions when using the 7B model as the retriever.}
\vspace{-0.4cm}
\scriptsize
\setlength{\tabcolsep}{2pt}
\renewcommand{\arraystretch}{1.1}
\resizebox{\columnwidth}{!}{%
  \begin{tabular}{@{}l c c c c c c@{}}
    \toprule
    \textbf{Method} & \textbf{VLLM} & \textbf{Size (B)} & \multicolumn{2}{c}{\textbf{Image retrieval}} & \multicolumn{2}{c}{\textbf{Text retrieval}} \\
    \cmidrule(lr){4-5}\cmidrule(lr){6-7}
    & & & R@1 & R@10 & R@1 & R@10 \\
    \midrule
    \lexilargezs & InternVL2.5-MPO & 4  & 75.2 & 95.7 & 81.0 & 97.4 \\
    \lexilargezs & InternVL2.5-MPO & 8  & 74.4 & 95.1 & \textbf{82.6} & \textbf{98.0} \\
    \lexilargezs & InternVL3       & 9  & \textbf{75.6} & 95.4 & 82.3 & 97.8 \\
    \lexilargezs & InternVL3       & 14 & 74.3 & \textbf{95.8} & 80.6 & 96.8 \\
    \bottomrule
  \end{tabular}%
}
\label{tab:ablate_size_of_vllm_7b_model_inference}
\end{table}

\subsection{SugarCrepe Detailed Results}

\vspace{1em}
\begin{table*}[ht]
\centering
\caption{Comparison with state-of-the-art on the SugarCrepe compositionality benchmark.}
\vspace{0.5em}
\scriptsize            
\setlength{\tabcolsep}{2pt}  
\renewcommand{\arraystretch}{1.1}
\resizebox{0.85\textwidth}{!}{
\begin{tabular}{@{}lccccccccccc@{}}
\toprule
\textbf{Method} & \textbf{Params} & \multicolumn{3}{c}{\textbf{Replace}} & \multicolumn{2}{c}{\textbf{Swap}} & \multicolumn{2}{c}{\textbf{Add}} & \textbf{Avg.} \\
& \textbf{(B)} & \textbf{Object} & \textbf{Attribute} & \textbf{Relation} & \textbf{Object} & \textbf{Attribute} & \textbf{Object} & \textbf{Attribute} & \\
\cmidrule(lr){1-2}
\cmidrule(lr){3-5}
\cmidrule(lr){6-7}
\cmidrule(lr){8-9}
\cmidrule(lr){10-10}
CLIP (ViT-B)~\cite{radford2021learning} & 0.15 & 90.9 & 80.1 & 69.2 & 61.4 & 64.0 & 77.2 & 68.8 & 73.1 \\
SigLIP ViT-B/16~\cite{zhai2023sigmoid} & 0.15 & 95.3 & 86.7 & 70.3 & 60.0 & 71.5 & 89.1 & 83.8 & 79.5 \\
CLIP (ViT-L)~\cite{radford2021learning} & 0.43 & 94.1 & 79.2 & 65.2 & 60.2 & 62.3 & 78.3 & 71.5 & 73.0 \\
EVA-02-CLIP (ViT-L-336)~\cite{fang2023eva02} & 0.43 & 96.6 & 85.1 & 70.9 & 64.9 & 65.3 & \textbf{92.9} & 82.1 & 80.2 \\
BLIP (ViT-L)~\cite{li2022blip}& 0.23 & 96.5 & 81.7 & 69.1 & 66.6 & 76.8 & 92.0 & 85.1 & 81.1 \\
BLIP2 (ViT-L)~\cite{li2023blip2} & 1.17 & \textbf{97.6} & 81.7 & 77.8 & 62.1 & 65.5 & 92.4 & 87.4 & 80.6 \\
OpenCLIP (ViT-G/14)~\cite{NEURIPS2022_a1859deb}  & 1.37 & 95.8 & 85.0 & 72.4 & 63.0 & 71.2 & 91.5 & 82.1 & 80.1 \\
OpenCLIP (ViT-BigG/14)~\cite{NEURIPS2022_a1859deb} & 2.54 & 96.6 & \textbf{87.9} & 74.9 & 62.5 & 75.2 & 92.2 & 84.5 & 81.9 \\
\midrule
NegCLIP~\cite{yuksekgonul2023visionlanguagemodelsbehavelike}& 0.15 & 92.7 & 85.9 & 76.5 & \textbf{75.2} & 75.4 & 88.8 & 82.8 & 82.5 \\
\midrule
\lexismallzs & 0.3 & 94.0 & 82.5 &\textbf{ 79.3} & 63.7 & 59.6 & 84.1 & 86.8 & 78.6 \\
\lexismallft{} & 0.3 & 96.7  & 86.3 & 77.5 & 69.0 &  \textbf{82.3}  & 90.7  & \textbf{91.9}  & \textbf{84.9} \\
\bottomrule

\end{tabular}
}
\vspace{-0.3cm}
\label{table:sugarcrepe_results_detailed}
\end{table*}

Table \ref{table:sugarcrepe_results_detailed} shows the detailed results of our method in all seven categories. Our finetuned model achieves state-of-the-art performance. 

\begin{table*}[!t]
  \centering
  \caption{Zero-shot text–image and image–text retrieval on \textbf{NoCaps} in the Out-of-Domain partition.}
  \small
  \setlength{\tabcolsep}{5pt}
  \renewcommand{\arraystretch}{1.05}
  \resizebox{\textwidth}{!}{%
    \begin{tabular}{@{}l c c c c c c c@{}}
      \toprule
      \textbf{Method} & \textbf{Params (B)}
        & \multicolumn{3}{c}{\textbf{Image retrieval}} 
        & \multicolumn{3}{c}{\textbf{Text retrieval}} \\
      \cmidrule(lr){3-5} \cmidrule(lr){6-8}
        &  & R@1 & R@5 & R@10 & R@1 & R@5 & R@10 \\
      \midrule
      CLIP (ViT-B/16)~\cite{radford2021learning}    
        & 0.15 & 53.5 & 85.7 & 93.3 & 69.4 & 93.6 & 97.8 \\
      CLIP (ViT-L/14)~\cite{radford2021learning}    
        & 0.43 & 56.8 & 87.0 & 93.2 & 74.9 & 95.8 & 98.4 \\
      OpenCLIP (ViT-G/14)~\cite{NEURIPS2022_a1859deb}  
        & 1.37 & 70.8 & 93.5 & 97.3 & 85.6 & 98.2 & \textbf{99.8} \\
      OpenCLIP (ViT-bigG/14)~\cite{NEURIPS2022_a1859deb}
        & 2.54 & \textbf{72.2} & 93.9 & 97.3 & 85.4 & \textbf{98.8} & 99.5 \\
      SigLIP ViT-B/16~\cite{zhai2023sigmoid}
        & 0.23 & 71.5 & 93.6 & 97.3 & 85.3 & 98.6 & 99.8 \\
      EVA-02-CLIP (ViT-L-336)~\cite{fang2023eva02}
        & 0.43 & 66.6 & 91.2 & 96.0 & 81.4 & 97.4 & 98.9 \\
      \midrule
      \lexismallzs
        & 0.30 & 67.8 & 91.4 & 96.5 & 79.4 & 96.4 & 98.5 \\
      \lexismallft{}
        & 0.30 & 71.0 & \textbf{93.7} & 97.3 & \textbf{85.9} & 97.3 & 99.3 \\
      \bottomrule
    \end{tabular}%
  }
  \label{tab:map_f1_nocaps}
  \vspace{-0.3cm}
\end{table*}

\begin{table*}[!t]
  \centering
  \caption{Zero-shot compositional retrieval on Winoground~\cite{diwan2022winogroundhardinvestigatingfailures} across Group, Image, and Text scores.}
  \small
  \setlength{\tabcolsep}{5pt}
  \renewcommand{\arraystretch}{1.05}
  \resizebox{0.7\textwidth}{!}{%
    \begin{tabular}{@{}l c c c c@{}}
      \toprule
      \textbf{Model} & \textbf{Params (B)} & \textbf{Image} & \textbf{Text} & \textbf{Group} \\
      \midrule
      CLIP (ViT-B/16)~\cite{radford2021learning} & 0.15 & 10.5 & 25.0 & 7.3 \\
      CLIP (ViT-L/14)~\cite{radford2021learning} & 0.43 & 12.3 & 27.5 & 8.3 \\
      OpenCLIP (ViT-G/14)~\cite{NEURIPS2022_a1859deb} & 1.37 & 12.8 & 32.0 & 9.3 \\
      OpenCLIP (ViT-bigG/14)~\cite{NEURIPS2022_a1859deb} & 2.54 & 15.5 & 35.5 & 12.0 \\
      SigLIP ViT-B/16~\cite{zhai2023sigmoid} & 0.23 & 13.0 & 33.0 & 10.5 \\
      \midrule
      \lexismallzs & 0.33 & 6.7 & 24.7 & 3.7 \\
      \lexismallft & 0.33 & 13.3 & 35.8 & 10.8 \\
      \bottomrule
    \end{tabular}%
  }
  \label{tab:winoground_results}
  \vspace{-0.3cm}
\end{table*}

\subsection{Generalisation to out-of-domain data}

To further validate our approach, we evaluated it on two additional datasets: (1)  Out-of-Domain Subset NoCaps~\cite{agrawal2019nocaps}, a subset specifically designed to assess retrieval performance on images that are out-of-domain relative to the COCO and Flickr datasets. (2) Winoground~\cite{diwan2022winogroundhardinvestigatingfailures}, a challenging dataset exhibiting a combination of unusual, adversarial images, sketches, etc.

As the results below demonstrate, our approach achieves performance comparable to significantly larger models like OpenCLIP 2.54B. This is noteworthy given that our model was trained on only 1.5 million textual samples, whereas OpenCLIP utilized a massive 5 billion image-text pairs.

\section{Ethics Statement and Artifacts}

\noindent \textbf{Licenses.} We follow the original licenses of all datasets and models used in this work. Our released artifacts (code, subFlickr, subCOCO) will be distributed under the MIT license.  

\noindent \textbf{Intended Use.} All datasets and models used in this work were employed strictly for research purposes, in accordance with their original intended use. Our derived benchmarks (subFlickr and subCOCO) are released exclusively for research use, consistent with the original licensing and access conditions of COCO and Flickr30k.

\noindent \textbf{Documentation.} All datasets and models used in this work are well-documented in their original publications.  Our derived benchmarks (subFlickr and subCOCO) contain short, compositional English queries paired with corresponding images, designed to evaluate fine-grained retrieval. We release the benchmarks with accompanying documentation to ensure transparency and reproducibility.

\end{document}